%% file: main.tex
\pdfoutput=1
\PassOptionsToPackage{table}{xcolor}
\documentclass{vgtc}

\graphicspath{{figures/}{pictures/}{images/}{./}}

\usepackage{times}

\usepackage{mathptmx}
\usepackage{amsmath}
\usepackage{amssymb}
\usepackage{booktabs}
\usepackage{array}
\usepackage{multirow}
\usepackage{url}
\usepackage{xcolor}
\usepackage{colortbl}
\usepackage{float}
\usepackage{tikz}
\usetikzlibrary{arrows.meta,positioning,calc,fit,shapes.geometric,decorations.pathreplacing}

\definecolor{improve}{HTML}{2A78D6}
\definecolor{degrade}{HTML}{D6452A}

\newif\ifshowtodos \showtodostrue


\newcommand{\Wi}{W_{2,\infty}}
\newcommand{\Wt}{W_{2,2}}
\newcommand{\Lcv}{L_{\mathrm{CV}}}
\newcommand{\Lpers}{L_{\mathrm{pers}}}
\newcommand{\Lmax}{L_{\mathrm{max}}}

\onlineid{0}
\nocopyrightspace
\vgtccategory{Research}
\vgtcinsertpkg

\title{Topology-Inspired Fine-Tuning for Wind-Field Super-Resolution}

\author{Arjun Dadhwal and Brian Summa}

\abstract{
Scientific super-resolution (SR) models are typically optimized with pointwise reconstruction losses, yet scientific interpretation often hinges on extrema, connected regions, and multiscale scalar structure. We characterize this tension with pretrained CNN and GAN wind-field SR models. Pointwise metrics, bottleneck persistence-diagram (PD) distance, and merge-tree (MT) distance favor the CNN, whereas order-2 Wasserstein PD distances favor the sharper GAN, so even different distances on the same topological descriptor reward different properties. Motivated by this disagreement, we fine-tune the pretrained CNN with differentiable topology-inspired auxiliary losses on scalar speed, gradient magnitude, soft high-speed level sets, and local maxima, and with fixed-pair supervision, which constrains predicted values and persistence at ground-truth persistence-pair endpoints located offline with the Topology ToolKit. A loss study of 16 fine-tuning configurations, each compared with a matched standard fine-tuning control that uses only the original reconstruction loss, on a fixed 168-field benchmark (with topology evaluated on a fixed $160\times160$ subregion of each field), shows that, among the scalar-field losses, gradient supervision provides the strongest PD improvement but does not improve mean MT distance, whereas fixed-pair supervision gives the largest and most consistent joint improvement in PD and MT distances. The final objective combines these two signals. Relative to the pretrained CNN, it lowers mean bottleneck, $W_{2,\infty}$, and $W_{2,2}$ distances by 31\%, 25\%, and 27\% (and by 32--35\% relative to the control, which on its own worsens all three), lowers all three on 150 of 168 fields, and gains 1.30~dB PSNR, about half the gain of standard fine-tuning. Mean MT distance also improves, but on only 91 of those 150 fields, indicating that PD and MT capture complementary structure.
}


\begin{document}

\firstsection{Introduction}
\maketitle

{\let\thefootnote\relax\footnote{Arjun Dadhwal and Brian Summa are with Tulane University. E-mail: adadhwal@tulane.edu, bsumma@tulane.edu. This article extends work accepted as a poster at TopoInVis Connect 2026. Relative to the reviewed workshop submission, it computes persistence-diagram distances with explicit implementations of the standard bottleneck and order-2 Wasserstein definitions, uses a corrected implementation of fixed-pair supervision, and adds the full loss study and additional validation.}}Scientific super-resolution (SR) is commonly optimized and evaluated using pointwise reconstruction fidelity, even though scientific interpretation often depends on connected regions, extrema, and structures that evolve across scalar thresholds. In wind-field applications, reconstructing such fine-scale structure supports renewable-energy assessment, extreme-event characterization, atmospheric analysis, and scientific visualization. However, many widely available datasets remain too coarse to resolve local ridges, high-speed regions, and connected structures directly~\cite{Stengel2020,Buster2024,Ding2025}.

Deep learning methods, including convolutional neural networks (CNNs) and generative adversarial networks (GANs), can substantially improve spatial resolution. Their evaluation remains difficult, however, because conventional measures such as peak signal-to-noise ratio (PSNR), mean absolute error (MAE), root-mean-square error (RMSE), and structural similarity (SSIM) primarily quantify pointwise or local reconstruction fidelity. These measures do not necessarily determine whether a reconstructed field preserves the scalar structures most relevant to scientific interpretation.

This limitation is related to the broader perception--distortion trade-off. CNNs trained primarily with reconstruction losses often produce conservative and comparatively smooth estimates with strong pointwise accuracy. Adversarial models can generate sharper and more spatially structured outputs, but they may also introduce unsupported fragmentation or visually plausible features absent from the ground truth. Blau and Michaeli~\cite{Blau2018} formalize the conflict between minimizing distortion and maximizing perceptual quality, while scientific-downscaling studies report corresponding disagreements between numerical-error rankings and visual, spatial-frequency, or distributional assessments. Stengel et al.~\cite{Stengel2020}, for example, observe that adversarial wind and solar reconstructions can appear sharper and more spatially detailed than CNN outputs even when pixelwise measures favor the CNN. H\"ohlein et al.~\cite{Hoehlein2020} likewise discuss limitations of MSE-based wind-field evaluation, noting that MSE can overweight high-speed magnitude errors while underweighting directional errors in low-speed regions. Annau and Cannon~\cite{Annau2023} further show that GAN-based wind downscaling can generate plausible fine-scale structures that need not coincide with the exact high-resolution reference, and that pixelwise metrics such as MAE/MSE can disagree with spatial-frequency and distributional measures of generated-field quality.

Topological descriptors provide a complementary means of examining such reconstructions. For a scalar field, a persistence diagram (PD) summarizes critical-point pairings and the scalar-value intervals over which corresponding features persist. A merge tree (MT) additionally represents how connected components evolve and merge across scalar thresholds. The two descriptors are therefore related but not interchangeable: persistence diagrams emphasize feature lifetimes, whereas merge trees preserve more of the connectivity and hierarchical organization of the scalar field~\cite{Yan2021}. Prior work has incorporated topological information into neural objectives for reconstruction, segmentation, representation learning, and scalar-field interpolation~\cite{Gabrielsson2020,Hu2019,Moor2020,Kissi2025} and has applied topology-guided analysis to wind-speed fields~\cite{Qin2023}. Nevertheless, topology-inspired supervision remains comparatively underexplored in scientific SR.

We investigate this question using pretrained CNN and GAN models from the Physics-Informed Resolution-Enhancing (PhIRE) framework for climatological wind and solar SR introduced by Stengel et al.~\cite{Stengel2020}. The original data contain horizontal wind components $u$ and $v$; for scalar topological analysis, we evaluate the corresponding wind-speed magnitude. We first compare the pretrained models on a fixed benchmark of 168 fields, held out from all fine-tuning, using a broad evaluation suite spanning direct fidelity, physics-oriented errors, spectral and distributional structure, and topology.

The baseline comparison reveals a descriptor-dependent trade-off. The CNN is preferred on PSNR for all 168 fields and on speed SSIM for 165 of 168 fields, and is generally stronger on direct wind-speed and wind-power-density errors. The GAN is more competitive on frequency-domain behavior, gradient distributions, and the most extreme exceedance measures, consistent with its tendency to produce sharper and more variable structure. The topological evaluators do not agree with one another either. MT distance favors the CNN on 148 of 168 fields, and bottleneck PD distance, which is governed by the single worst-matched feature, also favors the CNN (129 of 168). In contrast, the order-2 Wasserstein PD distances, which accumulate mismatch over all persistence pairs, favor the GAN on 155 and 158 fields. This pattern is consistent with the sharper GAN better matching the aggregate distribution of ground-truth birth--death pairs while incurring larger individual discrepancies and preserving the merge hierarchy less consistently. Neither conventional fidelity measures nor any single topological distance provides a complete account of reconstruction quality.

These observations motivate the central question of this work: can a fidelity-oriented CNN be fine-tuned toward persistence-relevant scalar structure while retaining strong conventional fidelity? We approach this question through a loss study rather than a single proposed loss. Starting from the pretrained CNN, we fine-tune with differentiable auxiliary losses that target different aspects of scalar structure: scalar speed, gradient magnitude, soft high-speed level sets, high-speed local maxima, and \emph{fixed-pair supervision}, which uses the Topology ToolKit (TTK)~\cite{Tierny2018} offline to locate the endpoints of ground-truth persistence pairs and penalizes errors in the predicted values and persistence at those fixed locations. None of these losses computes a persistence diagram or merge tree of the prediction; PD and MT distances are used only to evaluate the trained models.

Every configuration is compared with a matched control that is fine-tuned identically (same initialization, data, optimizer, and schedule) but with the reconstruction loss alone; we call it the standard fine-tuning control. Differences from this control therefore measure the effect of the auxiliary losses rather than that of continued fine-tuning. We trained and report 16 configurations in total, including all eight combinations of the speed, gradient, and level-set losses, which allows the contribution of each term and their interactions to be estimated. The study identifies gradient supervision and fixed-pair supervision as complementary signals, and we use their combination as the final objective.

The final objective lowers all three PD distances relative to both the pretrained CNN (by 25--31\% in the mean) and the matched control (by 32--35\%), whereas standard fine-tuning on its own makes all three worse; all three PD distances are lower than the CNN's on 150 of the 168 benchmark fields. The final model retains about half of the PSNR gain of standard fine-tuning. Mean MT distance also improves, but MT gets worse on 59 of the 150 fields whose PD distances all improve, indicating that PD and MT measure complementary information.

This work provides a controlled proof of concept for topology-inspired supervision in scientific SR. Its principal contributions are:

\begin{itemize}
\item a characterization of descriptor-dependent disagreement in wind-field SR, showing that pretrained CNN and GAN reconstructions are ranked differently by fidelity, physics-oriented, merge-tree, and even different persistence-diagram distances, which motivates topology-aware supervision of the fidelity-oriented CNN;

\item a family of lightweight, differentiable topology-inspired losses for fine-tuning an SR network, culminating in an objective that combines gradient-magnitude supervision with fixed-pair supervision at TTK-identified ground-truth persistence-pair endpoints;

\item a controlled comparison with fine-tuning on the reconstruction loss alone, showing that the persistence improvements are not explained by continued fine-tuning;

\item a fully reported loss study of 16 configurations with factorial analyses, showing that gradient supervision produces the largest PD improvement among the scalar-field losses but does not improve MT agreement, whereas fixed-pair supervision gives the largest and most consistent joint PD and MT improvement; and

\item an evaluation with three PD distances and MT distance, using a validated PD-distance implementation, showing that PD and MT improvements are frequently decoupled at the level of individual fields and therefore capture complementary information.
\end{itemize}

The remainder of the paper reviews wind-field SR and scalar-field topology, describes the data and controlled fine-tuning design, introduces the auxiliary losses and the loss study, and evaluates their effects using fidelity, physics-oriented, PD, and MT measures.

\section{Background and Related Work}

\subsection{Wind-field super-resolution and evaluation}
Stengel et al.~\cite{Stengel2020} introduced the PhIRE framework for large-factor SR of climatological wind and solar data using CNN and GAN models. Subsequent work has explored transformer, diffusion, and generative approaches for wind-field downscaling and renewable-energy applications~\cite{Tang2024,Merizzi2024,Buster2024,Ding2025}. Complementary work on constrained climate downscaling has shown that physical consistency constraints can be built into neural SR models, motivating evaluation beyond visual or pointwise fidelity~\cite{harder2024hardconstraineddeeplearningclimate}. At the same time, prior studies have documented disagreement between pointwise numerical errors and visual, spatial-frequency, or distributional assessments~\cite{Hoehlein2020,Annau2023}. Such disagreement is especially consequential for scientific fields, where sharper output can represent either recovered physical structure or model-induced artifacts. We therefore evaluate not only PSNR and SSIM, but also wind-power-density (WPD), spectral, gradient, exceedance, component-structure, and topological measures.

Beyond meteorological downscaling, super-resolution has been studied directly for scientific vector-field visualization. Here, spatial SR estimates a higher-resolution field from a coarser one at the same time instant, whereas temporal SR reconstructs missing intermediate time steps. Guo et al.~\cite{Guo2020SSRVFD} proposed SSR-VFD for spatial SR of vector fields, An et al.~\cite{An2021STSRNet} joint space--time SR, and Han and Wang~\cite{Han2022TSRVFD} temporal SR of unsteady vector fields; these methods aim to reconstruct vector-valued data in a form useful for downstream visualization and analysis. Our setting is complementary. The SR network still reconstructs the complete wind vector $[u,v]$, and topology-inspired supervision is applied to the derived scalar wind speed in order to study persistence and merge hierarchy.

\subsection{Scalar filtrations and topological descriptors}
\label{sec:topology-background}
Let $s:\Omega\rightarrow\mathbb{R}$ denote a scalar wind-speed field on a spatial domain $\Omega$. Its topology can be studied by sweeping a scalar threshold through the field. The sublevel and superlevel sets are
\begin{equation}
    S^-_{\tau}=\{x\in\Omega\mid s(x)\leq\tau\},
    \qquad
    S^+_{\tau}=\{x\in\Omega\mid s(x)\geq\tau\}.
\end{equation}
As $\tau$ changes, connected components appear and merge only at critical values. Sublevel filtrations emphasize low-valued basins; superlevel filtrations provide the dual view of high-valued peaks and regions. Several of our auxiliary losses target high-speed regions and local maxima, which are most intuitive from the superlevel view, in which a prominent maximum gives birth to a high-speed component. All reported topological descriptors, and the persistence pairs used by fixed-pair supervision, instead use the sublevel convention of our TTK pipeline. The two views are dual: in a sublevel filtration of a two-dimensional field, a local maximum does not create a component but closes a cycle~\cite{EdelsbrunnerHarer2010}. Under our sublevel convention, maxima therefore appear as death endpoints of finite saddle--maximum pairs ($D_1$), one of the two pair families used in our evaluation (Sec.~\ref{sec:evaluation}). The high-speed extrema targeted by our local-maximum loss thus overlap conceptually with persistence-relevant critical values, although not every maximum ends a long-lived pair.

\paragraph{Persistence diagrams.}
A persistence diagram records birth--death pairs
\begin{equation}
    D(s)=\{(b_i,d_i)\}_{i=1}^{n},
\end{equation}
where the persistence $|d_i-b_i|$ measures the scalar interval over which a feature remains present. Points close to the diagonal $b=d$ represent short-lived features, whereas points farther from it represent more persistent structure. PDs compactly preserve paired critical values but not the complete ancestry describing which components merge into one another, nor where in the domain a feature lies.

We compare diagrams with three distances. The bottleneck distance
\begin{equation}
 d_B(D_1,D_2)=
 \inf_{\gamma}\sup_{p\in D_1}\lVert p-\gamma(p)\rVert_{\infty},
 \label{eq:bottleneck}
\end{equation}
where $\gamma$ ranges over matchings between diagram points, including matches to the diagonal, is controlled by the single largest discrepancy under an optimal matching. The order-2 Wasserstein distance with ground norm $\lVert\cdot\rVert_q$,
\begin{equation}
 W_{2,q}(D_1,D_2)=
 \Bigl(\inf_{\gamma}\sum_{p\in D_1}\lVert p-\gamma(p)\rVert_{q}^{2}\Bigr)^{1/2},
 \qquad q\in\{\infty,2\},
 \label{eq:wasserstein}
\end{equation}
instead accumulates squared discrepancies over all matched pairs. We report both $\Wi$ ($L_\infty$ ground norm, matching the bottleneck geometry) and $\Wt$ (Euclidean ground norm); the two differ only in the ground norm, not in the outer order. Because ``$W_2$'' is used for both conventions in the literature, we always write both indices. Because $d_B$ measures the worst-case mismatch and $W_{2,q}$ the accumulated mismatch, they can rank reconstructions differently: a field that matches the bulk of the diagram well but misplaces one prominent feature can have a low Wasserstein and a high bottleneck distance, and vice versa.

\paragraph{Merge trees.}
A merge tree retains the connected-component evolution together with the hierarchy of merge events. Leaves represent extrema at which components are created, internal nodes represent saddles at which components merge, and edges encode scalar intervals between events. Consequently, an MT can distinguish fields that have similar birth--death pairs but different ancestry, branch order, or connectivity~\cite{Yan2021,PontPGA_2022,PontW2_2021}.

\begin{table}[t]
\caption{Complementary information retained by the two topological descriptors used in this study.}
\label{tab:descriptor-comparison}
\centering
\scriptsize
\begin{tabular}{p{0.16\columnwidth}p{0.35\columnwidth}p{0.35\columnwidth}}
\toprule
Descriptor & Retains & Does not explicitly retain\\
\midrule
PD & Paired birth/death values and feature persistence & Full component ancestry and branch hierarchy; spatial location of features\\
MT & Critical values, component ancestry, and merge hierarchy & Original spatial embedding of the complete field\\
\bottomrule
\end{tabular}
\end{table}

\input{figures/topology_descriptor_schematic_tikz.tex}

Recent work emphasizes that descriptor choice and distance choice both affect scalar-field comparison. Wasserstein formulations support merge-tree geodesics, barycenters, and representation learning~\cite{PontW2_2021,PontWAE_2023,WetzelsPath_2023}; complementary approaches study stable, geometry-aware, branch-independent, and deformation-based comparisons~\cite{Bollen_2023,YanGeometryAware_2021,WetzelsBranch_2022,WetzelsDeformation_2022}. These developments motivate treating PD and MT evaluations, and different distances on the same descriptor, as related but distinct evidence rather than interchangeable notions of a single ``topological quality.''

\subsection{Topology in neural learning}
\label{sec:related-learning}
Persistent topology has been incorporated into neural models through differentiable layers, regularizers, and task-specific losses. Br\"uel-Gabrielsson et al.~\cite{Gabrielsson2020} demonstrate differentiation through persistence for reconstruction and generative-model regularization, while Hu et al.~\cite{Hu2019} and Clough et al.~\cite{Clough_2022} use differentiable topological losses to reduce connectivity or shape errors in image and volume segmentation. Moor et al.~\cite{Moor2020} preserve multiscale connectivity in autoencoder latent spaces, and Kazeminia et al.~\cite{Kazeminia2026TGMIL} use a topological inductive bias to improve multiple-instance learning under data-scarce conditions. Explicit optimization has also been investigated for scalar-data simplification~\cite{KissiSimplification_2024}, and efficient persistence computation is an active topic in scientific visualization~\cite{GuillouDMS_2023}. Learned representations have been proposed for efficient merge-tree comparison; Qin et al.~\cite{QinMTNN2024} approximate MT similarity with graph neural networks and topological attention.

The critical-value component of our fixed-pair supervision is closely related to the topology-aware interpolation framework of Kissi et al.~\cite{Kissi2025}. Their training proceeds in two phases. The first combines reconstruction and gradient losses with a critical-value loss that constrains predicted values at the birth and death vertices of the target persistence pairs identified by TTK. The second keeps the reconstruction and critical-value terms, drops the gradient term, and adds a Wasserstein-type persistence-diagram loss: at each iteration the prediction's diagram is recomputed with TTK and matched to the target diagram, prediction pairs without a counterpart are pushed toward the diagonal, and the matched birth--death cost is differentiated with respect to the predicted values at the prediction's critical vertices. Our critical-value term $\Lcv$ closely follows their first-phase constraint. We differ in three respects. First, we add a persistence term $\Lpers$ that directly constrains the scalar contrast between the two endpoints of each supervised pair. Second, we do not recompute or re-match the prediction's persistence diagram during training: supervision is confined to fixed ground-truth endpoint locations and, unlike their second-phase loss, cannot remove spurious pairs that the prediction creates elsewhere. Third, we fine-tune a pretrained wind-field SR network rather than train a time-conditioned scalar-field decoder. The PD and MT distances we report are therefore evaluation measures that our training procedure never computes.

A related line of work preserves topological and other features during lossy compression, for scalar fields~\cite{Yan2024TopoSZ}, vector fields~\cite{Liang2023FeatureVF}, and as a general augmentation of existing compressors~\cite{Gorski2025TopoA}. These methods share our motivation of retaining analysis-relevant structure under an information-reducing transformation, but the problems differ: a compressor has the original high-resolution field at encoding time and can enforce guarantees on the decoded field, whereas SR receives only a coarse observation and must infer unresolved structure. We therefore treat topology-preserving compression as methodological context rather than a directly comparable baseline.

\section{Data and Experimental Setup}
\label{sec:data-setup}

\subsection{Data source and PhIRE-compatible preprocessing}
Wind fields are drawn from the NREL WIND Toolkit national gridded dataset (NREL, now the National Laboratory of the Rockies, NLR) and accessed through the Highly Scalable Data Service (HSDS). The upstream variables are 100~m wind speed and wind direction, which are converted to horizontal vector components $[u,v]$ using the standard meteorological convention. We use a Mid-Atlantic spatial crop centered near $39.5^{\circ}$N, $75.0^{\circ}$W and follow the PhIRE wind-data organization in which raw high-resolution vector fields are downsampled to form paired lower-resolution inputs and higher-resolution targets. The PhIRE data hierarchy contains a low-resolution to medium-resolution (LR$\rightarrow$MR) stage and an MR$\rightarrow$HR stage. We restrict this study to MR$\rightarrow$HR so that all loss configurations share a single pretrained model, resolution factor, and evaluation pipeline; the losses are not specific to this stage and could in principle be applied to LR$\rightarrow$MR, but we have not evaluated that setting.

The benchmark examples are represented as two-channel vector fields. Each medium-resolution (MR) input is a $100\times100\times2$ field paired with a $500\times500\times2$ high-resolution (HR) ground-truth (GT) field. The SR networks operate on normalized $u$ and $v$ components, while topology and most domain-specific analyses are computed after denormalization on scalar wind-speed magnitude $s=\sqrt{u^2+v^2}$. Every method is evaluated against the same benchmark ground truth.

For topological analysis, we use a fixed $160\times160$ top-left crop of each $500\times500$ denormalized speed field (about 10\% of its area). This crop is used consistently for persistence-diagram extraction, merge-tree evaluation, and the ground-truth persistence pairs used by fixed-pair supervision. It was introduced for computational tractability, not because the subregion has special physical significance. Conventional and domain-specific metrics are evaluated on the full reconstructed fields.

\subsection{Evaluation benchmark and fine-tuning set}
\label{sec:benchmark}
The fixed evaluation benchmark constructed for this study contains 168 consecutive hourly fields from the same spatial region. It is used for the pretrained CNN/GAN comparison and for every fine-tuned model, so all reported winner counts and mean metrics are computed on the same 168 fields. Because the benchmark is also used to compare loss configurations (Sec.~\ref{sec:selection}), it serves as both the evaluation set and the set on which the final configuration was chosen; there is no separate validation split.

For fine-tuning, we build a separate set of 2688 samples using the same spatial region, variable conversion, normalization convention, and MR$\rightarrow$HR preprocessing. It consists of sixteen non-overlapping 168-hour windows, four per season, that are temporally disjoint from the evaluation benchmark. Every fine-tuned model reported in this paper is trained on this set.

\input{figures/data_pipeline_schematic_tikz.tex}

\subsection{Baselines, matched control, and loss configurations}
The pretrained PhIRE CNN and GAN provide fixed baselines. We use the released MR$\rightarrow$HR wind models of Stengel et al.~\cite{Stengel2020}, whose generator is a fully convolutional SRGAN-derived architecture with 16 residual blocks and skip connections; batch-normalization layers are omitted, and the wind network uses two input and output channels for the horizontal wind components. The CNN baseline is this generator after content-only pretraining with a pixelwise MSE loss; GAN training then continues from the pretrained generator with an additional discriminator and adversarial loss. We use the released checkpoints without retraining them. Every fine-tuned model starts from the pretrained CNN checkpoint and is fine-tuned for three epochs with Adam, learning rate $10^{-5}$, and batch size four, on the same 2688-field training set with the same normalization. All 16 configurations fine-tune the native PhIRE (TensorFlow) CNN generator itself; earlier experiments with separate PyTorch residual refiners were development studies and are excluded from this comparison. The \emph{standard fine-tuning control} optimizes only the original vector reconstruction loss $L_{uv}$. The final topology-inspired objective and the 14 other loss configurations add auxiliary losses to $L_{uv}$ and are otherwise identical to the control. Differences between any configuration and the control therefore isolate the effect of its auxiliary losses from that of continued fine-tuning and additional data.

\subsection{Reproducibility}
The source code, experiment scripts, evaluation utilities, and configuration files for this study are available in the tagged release \url{https://github.com/Ardad2/PhIRE/releases/tag/topoinvis-2026-arxiv-v1.1}, together with a reproducibility guide that maps the reported results to their code and data. The repository includes the data-processing path, the fine-tuning implementation for all configurations, and the scripts used to compute the reported fidelity, physics-oriented, distributional, and topological evaluations. The quantitative topology tables and loss-study figures are generated by script from a frozen table of per-field distances (18 methods $\times$ 168 fields), which is released with the code and can be used to check them independently. The qualitative Sample~78 visualization is generated separately from reconstructed fields and persistence diagrams, and the conceptual diagrams are authored with the manuscript source. Retraining the models and re-extracting the topological descriptors from scratch require additional computation and intermediate data that are regenerated by the released scripts.

\section{Topology-Inspired Fine-Tuning}
\label{sec:method}

\subsection{Vector reconstruction and scalar wind speed}
PhIRE super-resolves the horizontal wind vector. The network receives a coarse two-channel field of normalized $u$ (eastward) and $v$ (northward) components and predicts the same two components at five times the spatial resolution. Its original reconstruction loss $L_{uv}$ is the mean squared error between the predicted and ground-truth normalized components. Because this error is computed jointly on both components, it penalizes the complete vector error and is sensitive to discrepancies in both magnitude and orientation, although it contains no separate angular term. The standard fine-tuning control is therefore not an orientation-only model. Wind speed is the magnitude of this vector. All auxiliary losses are computed after denormalization on
\begin{equation}
 s(x,y)=\sqrt{u(x,y)^2+v(x,y)^2},
\end{equation}
which is also the scalar field used for PD, MT, and most domain metrics. We analyze scalar wind speed rather than vector-field topology because the persistence-diagram and merge-tree constructions used in this study are scalar-field descriptors and wind speed is a physically interpretable quantity, central to wind-energy applications. The analysis therefore does not assess vector-field topology, such as flow critical points or separatrices, which is a distinct direction.

\subsection{Why differentiable proxies}
\label{sec:why-proxies}
Directly optimizing persistence-diagram agreement requires constructing the prediction's diagram and matching it to the target during training. Such a procedure can be made piecewise differentiable by holding the current pairing and matching fixed and differentiating the matched cost with respect to the predicted values at critical vertices, as in prior work~\cite{Gabrielsson2020,Kissi2025}. In our setting, persistence diagrams are extracted by an external TTK computation that is not part of the PhIRE training graph, so this approach would require extracting and matching a diagram for every prediction at every training step. We instead use auxiliary losses that are ordinary differentiable functions of the predicted speed field; topology enters training only through ground-truth locations that are computed once, before training. This keeps fine-tuning inside the existing training loop, but it also means that none of our losses optimizes $d_B$, $W_{2,q}$, or MT distance directly. Agreement under those distances is an empirical outcome that we measure.

We also piloted a more direct formulation: a differentiable cubical-complex persistence-diagram Wasserstein loss applied inside a small residual refiner. It made persistence supervision active during training but did not improve the final TTK-based PD or MT distances on the evaluation topology crop. Because that pilot differed from the final evaluation in crop size, model capacity, implementation framework, and diagram construction, we interpret it cautiously: it suggests that persistence-oriented losses need to be aligned with the evaluated scalar field and diagram convention, not that differentiable topology is ineffective. Fixed-pair supervision (Sec.~\ref{sec:loss-components}) satisfies this alignment by construction, since its pairs come from the same TTK convention and the same fixed $160\times160$ topology crop used for evaluation.

\subsection{Topology-inspired loss components}
\label{sec:loss-components}
\label{sec:ablation-terms}
We developed five auxiliary losses in roughly increasing order of topological specificity. The first three supervise the scalar field as a whole; the last two supervise values at locations selected from the ground-truth topology.

\textbf{Scalar speed.}
$L_{\mathrm{speed}}=\operatorname{MSE}(s_{\mathrm{SR}},s_{\mathrm{GT}})$ supervises the physical magnitude used by the topology evaluation, which $L_{uv}$ constrains only indirectly through the normalized components.

\textbf{Gradient magnitude.}
\begin{equation}
L_{\mathrm{grad}}=\operatorname{MSE}(|\nabla s_{\mathrm{SR}}|,|\nabla s_{\mathrm{GT}}|)
\end{equation}
penalizes errors in local ridges, fronts, and sharp transitions, counteracting the smoothing tendency of standard fine-tuning.

\textbf{Soft high-speed level sets.}
At thresholds $\mathcal{T}=\{5,10,15\}$~m/s, we define a differentiable occupancy mask $M_{\tau}(s)=\sigma\!\left(k(s-\tau)\right)$ with $k=10$, and compare predicted and ground-truth masks:
\begin{equation}
L_{\mathrm{levelset}}=\frac{1}{|\mathcal{T}|}\sum_{\tau\in\mathcal{T}}
\operatorname{MSE}\!\left(M_{\tau}(s_{\mathrm{SR}}),M_{\tau}(s_{\mathrm{GT}})\right).
\end{equation}
This term encourages agreement in the location and extent of high-speed regions across physically meaningful thresholds.

\textbf{High-speed local maxima.}
Our first attempt to target persistence-relevant critical values directly supervises prominent maxima without computing any persistence. For each ground-truth field, we select pixels that are local maxima within a $3\times3$ neighborhood and satisfy $s_{\mathrm{GT}}(x,y)\geq \mu_s+\sigma_s$, where $\mu_s$ and $\sigma_s$ are the spatial mean and standard deviation of the ground-truth speed for that sample. This per-field threshold is a simple heuristic for keeping maxima that stand out from the field's typical speed; it was not tuned. With $M_{\max}^{(b)}$ the resulting binary mask for batch element $b$,
\begin{equation}
\Lmax=
\frac{\sum_{b,x,y}M_{\max}^{(b)}(x,y)
\left(s_{\mathrm{SR}}^{(b)}(x,y)-s_{\mathrm{GT}}^{(b)}(x,y)\right)^2}
{\max\!\left(\sum_{b,x,y}M_{\max}^{(b)}(x,y),1\right)}.
\label{eq:lcrit}
\end{equation}
Under the sublevel convention these maxima are high critical values that end persistence pairs (Sec.~\ref{sec:topology-background}), but the term supervises neither the other endpoint of any pair nor its persistence. The $3\times3$ (eight-neighbor) test is also simpler than TTK's notion of a maximum: TTK uses an implicit triangulation of the regular grid together with a symbolic tie-breaking order for equal scalar values~\cite{Tierny2018}, whereas the pixel test can flag adjacent pixels on a plateau or identify a maximum that participates differently in TTK's pairing.

\textbf{Fixed-pair supervision.}
Fixed-pair supervision generalizes the local-maximum term by supervising both endpoints of actual ground-truth persistence pairs. For each ground-truth training field, TTK computes the persistence diagram of $s_{\mathrm{GT}}$ once, offline, on the same $160\times160$ topology crop and with the same convention as the evaluation. We consider finite pairs of both families ($D_0$ and $D_1$) with positive persistence, discard pairs whose persistence is below 1\% of the field's maximum persistence, rank the remainder by persistence, and retain the 64 most persistent; every field in the 2688-field training set supplies exactly 64 pairs. For each supervised pair $i$ we store the vertex $v^b_i$ at which the feature is born, the vertex $v^d_i$ at which it dies, and the ground-truth values $b_i=s_{\mathrm{GT}}(v^b_i)$, $d_i=s_{\mathrm{GT}}(v^d_i)$, and $p_i=|d_i-b_i|$. During training, the prediction is read at those fixed vertices, $\hat b_i=s_{\mathrm{SR}}(v^b_i)$ and $\hat d_i=s_{\mathrm{SR}}(v^d_i)$, and two terms are computed:
\begin{align}
\Lcv &= \tfrac12\bigl[\operatorname{MSE}(\hat b,b)+\operatorname{MSE}(\hat d,d)\bigr],\\
\Lpers &= \operatorname{MSE}(\hat p,p), \qquad \hat p_i=|\hat d_i-\hat b_i|,
\end{align}
where each MSE averages over all supervised pairs in the mini-batch; because every field contributes 64 pairs, fields are weighted equally. The critical-value term $\Lcv$ constrains the scalar values at both endpoints of each ground-truth pair, which include saddle-side as well as extremal critical points; it closely follows the critical-value constraint of Kissi et al.~\cite{Kissi2025}. The persistence term $\Lpers$ constrains the scalar contrast across each pair, which is the quantity that determines a feature's distance from the PD diagonal. Figure~\ref{fig:objective-schematic} summarizes the construction.

Fixed-pair supervision is a proxy, not a persistence-diagram loss. No diagram of the prediction is computed during training, the stored pairs are never re-matched to the prediction's own critical points, and nothing in the loss prevents the prediction from creating additional persistence pairs at other locations.

\input{figures/fixed_pair_objective_tikz.tex}

\subsection{Loss study design}
\label{sec:study-design}
Each configuration adds a subset of these losses to $L_{uv}$ with fixed weights: $0.01\,L_{\mathrm{speed}}$, $0.05\,L_{\mathrm{grad}}$, $0.25\,L_{\mathrm{levelset}}$, $0.001\,\Lmax$, and, for fixed-pair supervision, $0.004\,\Lcv+0.002\,\Lpers$. In tables and figures we name configurations by the losses they add: speed (S), gradient (G), level-set (L), local maximum (M), and fixed-pair (FP). The 16 configurations are the standard fine-tuning control; the seven other combinations of S, G, and L; M added to the control, to G, and to S+G+L; and FP added to the control, to G, to G+L, to S+G+L, and to S+G+L+M.

Because all eight combinations of S, G, and L were trained, the average effect of each of these losses, and their interactions, can be estimated (a $2^3$ factorial design). Two further $2\times2$ designs cross FP with M (both added to S+G+L) and with L (both added to G). All 16 configurations were trained once each at the 2688-sample scale, and we report every one.

\textbf{Loss weights.}
Coefficients were calibrated against the numerical scale of the unweighted losses and should not be read as measures of importance (Appendix~\ref{app:weights}). For example, the raw local-maximum loss was about $191\times$ larger than $L_{uv}$ in a calibration diagnostic at the pretrained checkpoint, so a coefficient of $0.001$ still contributed about 19\% of the reconstruction-loss magnitude; the speed, gradient, and level-set terms contributed about 40\%, 91\%, and 27\%. We chose coefficients that kept the auxiliary terms on a broadly comparable scale without severely degrading reconstruction fidelity. During development, a residual-refiner experiment with fixed-pair coefficients of $0.04$ and $0.02$ degraded fidelity (PSNR 30.56~dB versus 31.19~dB for the pretrained CNN), so both coefficients were reduced tenfold before the final experiments. No systematic hyperparameter search was performed. The reported coefficients are therefore not intended as universal constants. For a new dataset or architecture, we would measure the unweighted loss magnitudes at the pretrained checkpoint on a representative calibration set, choose coefficients that place the weighted auxiliary terms on a scale comparable to the reconstruction loss, and select among those settings on a separate validation set; level-set thresholds and pair-selection parameters should likewise be adapted to the scalar range of the target data.

\subsection{Final objective}
\label{sec:final-objective}
The loss study (Secs.~\ref{sec:ablation}--\ref{sec:selection}) identifies gradient supervision and fixed-pair supervision as complementary signals. The final objective combines them:
\begin{equation}
L = L_{uv}+0.05\,L_{\mathrm{grad}}+0.004\,\Lcv+0.002\,\Lpers .
\label{eq:f1-objective}
\end{equation}
Gradient supervision counteracts the smoothing of standard fine-tuning and gives the largest PD improvement among the scalar-field losses; fixed-pair supervision constrains values and persistence at ground-truth pair endpoints and provides the strongest joint PD and MT improvement. In tables and figures this configuration is G+FP.

\section{Evaluation Measures}
\label{sec:evaluation}
We report four groups of measures. Direct fidelity includes PSNR on the two denormalized wind components $[u,v]$, speed MAE/RMSE, and SSIM on scalar speed. For PSNR, the mean squared error is computed over both denormalized components, with the peak range defined by the minimum and maximum ground-truth component values of each field. Speed SSIM is computed on scalar wind-speed magnitude with a per-field data range spanning both the ground truth and the reconstruction. Physics-oriented metrics, distributional measures, component curves, and topological descriptors are computed on scalar wind-speed magnitude derived from the denormalized vector components. Physics and distributional measures include WPD errors and Wasserstein distance, power-spectral-density (PSD) errors, gradient errors and distributional statistics, and high-speed exceedance error. A component-curve $L_1$ measure compares connected-component counts over a dense threshold sweep and serves as a topology-adjacent structural proxy.

\textbf{Topology.}
For each reconstruction, TTK extracts the sublevel persistence diagram of the $160\times160$ speed crop. We keep the two finite pair families separately, $D_0$ (minimum--saddle) and $D_1$ (saddle--maximum), and exclude the single non-finite global pair. With our own implementation of Eqs.~\ref{eq:bottleneck}--\ref{eq:wasserstein}, we compute $d_B$, $\Wi$, and $\Wt$ between the reconstruction and ground-truth diagrams separately for $D_0$ and $D_1$, with unmatched points matched to the diagonal, and combine them as
\begin{equation}
 d_B=\max\bigl(d_B^{D_0},d_B^{D_1}\bigr),\qquad
 W_{2,q}=\Bigl[\bigl(W_{2,q}^{D_0}\bigr)^2+\bigl(W_{2,q}^{D_1}\bigr)^2\Bigr]^{1/2}.
 \label{eq:combine}
\end{equation}
This is equivalent to an optimal matching on the pooled finite diagram that pairs points only within the same family. Implementing the definitions explicitly avoids ambiguity from software-specific distance conventions. MT distance is computed on the same crops with TTK's merge-tree distance between join trees (merge trees of sublevel sets); for readability we call it MT distance throughout. All four distances are lower-is-better. Persistence thresholding is used only to reduce clutter in qualitative figures.

\textbf{Validation of the PD distances.}
Because the PD distances are our own implementation, we validated them along two independent paths. First, GUDHI's distance functions, applied to the same TTK-extracted $D_0$ and $D_1$ diagrams across 51 archived reconstruction runs and 168 fields per run (8568 ground-truth--reconstruction comparisons), reproduce our bottleneck distances exactly; the maximum absolute discrepancy is $1.02\times10^{-12}$ for $\Wi$ and $6.40\times10^{-14}$ for $\Wt$. This validates the distance computation rather than TTK's diagram construction. Second, to test sensitivity to the diagram-construction backend, we rebuilt the diagrams of the pretrained CNN, the standard fine-tuning control, and the final model with GUDHI's cubical complexes for all 168 fields. These diagrams are not identical to TTK's, because the two libraries use different discretization and pairing conventions, but they preserve our conclusions: the ranking of the three models agrees on 161, 165, and 164 of 168 fields for $d_B$, $\Wi$, and $\Wt$, and the direction of the final-model-versus-CNN comparison agrees on 98.8\%, 100\%, and 99.4\% of fields. For merge trees, an independent join-tree construction reproduces TTK's trees, after root normalization, for all 504 SR trees examined. This validates tree construction but not TTK's merge-tree distance itself, which we did not re-implement.

\textbf{Win counts.}
When we say that one method wins on a field, it has the higher PSNR/SSIM or the lower error or distance on that field; win counts are out of the 168 benchmark fields. We report means in the main text; medians (Appendix~\ref{app:medians}) lead to the same conclusions.

\section{Results}

\subsection{Pretrained CNN--GAN trade-off}
The pretrained models expose the evaluation tension that motivates this work. Table~\ref{tab:baseline-winners} reports per-field win counts for the complete evaluation suite. The CNN dominates direct errors, WPD MAE/RMSE, and gradient MAE. The GAN is preferred on PSD log-$L_2$, gradient Wasserstein distance, and slightly more often at the p95 and p99 exceedance thresholds.

The topology rows do not reduce to a single verdict either. MT distance prefers the CNN (148 of 168 fields), and so does bottleneck PD distance (129 of 168; mean $d_B$ 3.124 versus 5.059), whereas both Wasserstein PD distances prefer the GAN (155 and 158 of 168; mean $\Wt$ 17.46 versus 24.55). One reading, consistent with the GAN's sharper output and its behavior on the gradient and spectral measures, is that the GAN better matches the aggregate distribution of birth--death pairs that the Wasserstein distances accumulate, while producing larger worst-case discrepancies and a less faithful merge hierarchy. We offer this as an interpretation rather than an established mechanism. Either way, the preferred reconstruction depends on whether the evaluation emphasizes pointwise error, distributional sharpness, accumulated or worst-case persistence mismatch, or merge hierarchy.

\begin{table*}[t]
\caption{Per-field win counts for the pretrained CNN and GAN on the 168-field benchmark. Higher is better for PSNR and SSIM; lower is better for all listed errors and distances. No ties occurred in these comparisons.}
\label{tab:baseline-winners}
\centering
\scriptsize
\begin{tabular}{llcrr}
\toprule
Family & Metric & Prefer & CNN wins & GAN wins\\
\midrule
Direct & PSNR-$uv$ & $\uparrow$ & 168 & 0\\
Direct & SSIM-speed & $\uparrow$ & 165 & 3\\
\midrule
Wind power & WPD bias $|\cdot|$ & $\downarrow$ & 115 & 53\\
Wind power & WPD MAE & $\downarrow$ & 168 & 0\\
Wind power & WPD RMSE & $\downarrow$ & 167 & 1\\
Wind power & WPD W1 & $\downarrow$ & 119 & 49\\
\midrule
Spectrum & PSD log-$L_2$ & $\downarrow$ & 25 & 143\\
Spectrum & PSD slope $|\Delta|$ & $\downarrow$ & 77 & 91\\
\midrule
Gradients & Gradient MAE & $\downarrow$ & 168 & 0\\
Gradients & Gradient W1 & $\downarrow$ & 1 & 167\\
Gradients & Kurtosis $|\Delta|$ & $\downarrow$ & 96 & 72\\
\midrule
Extremes & Exceedance $|\Delta|$, $s>5$ & $\downarrow$ & 112 & 56\\
Extremes & Exceedance $|\Delta|$, $s>10$ & $\downarrow$ & 142 & 26\\
Extremes & Exceedance $|\Delta|$, $s>15$ & $\downarrow$ & 101 & 67\\
Extremes & Exceedance $|\Delta|$, p90 & $\downarrow$ & 104 & 64\\
Extremes & Exceedance $|\Delta|$, p95 & $\downarrow$ & 83 & 85\\
Extremes & Exceedance $|\Delta|$, p99 & $\downarrow$ & 76 & 92\\
\midrule
Topology & Bottleneck PD $d_B$ & $\downarrow$ & 129 & 39\\
Topology & Wasserstein PD $\Wi$ & $\downarrow$ & 13 & 155\\
Topology & Wasserstein PD $\Wt$ & $\downarrow$ & 10 & 158\\
Topology & Merge tree & $\downarrow$ & 148 & 20\\
\bottomrule
\end{tabular}
\end{table*}

\begin{table*}[t]
\caption{Final objective (G+FP) versus the pretrained CNN and the standard fine-tuning control on the 168-field benchmark. Means are shown; percentages are relative reductions in mean distance (positive: lower, i.e., better). ``Control vs.\ CNN'' shows the change from fine-tuning with the reconstruction loss alone; ``Final vs.\ control'' the change from adding the auxiliary losses under otherwise identical fine-tuning. The last two columns count the fields on which the final model has the lower distance.}
\label{tab:headline}
\centering
\scriptsize
\begin{tabular}{lrrrrrrrr}
\toprule
 & \multicolumn{3}{c}{Mean distance} & \multicolumn{3}{c}{Reduction in mean distance} & \multicolumn{2}{c}{Final lower than} \\
\cmidrule(lr){2-4}\cmidrule(lr){5-7}\cmidrule(lr){8-9}
Distance & CNN & Control & Final & Control vs.\ CNN & Final vs.\ CNN & Final vs.\ control & CNN & Control\\
\midrule
\input{tables/tab_headline.tex} 
\bottomrule
\end{tabular}
\end{table*}

\subsection{Persistence gains beyond ordinary fine-tuning}
\label{sec:headline}
Table~\ref{tab:headline} compares the final objective with the pretrained CNN and the standard fine-tuning control. Fine-tuning with the reconstruction loss alone makes persistence agreement \emph{worse}: the control increases all three mean PD distances relative to the pretrained CNN, by 5.2\% ($d_B$), 10.0\% ($\Wi$), and 12.0\% ($\Wt$), and lowers all three on only 11 of 168 fields. Adding the auxiliary losses reverses this. The final model lowers mean $d_B$, $\Wi$, and $\Wt$ by 31.2\%, 25.2\%, and 27.4\% relative to the CNN and by 34.6\%, 32.0\%, and 35.2\% relative to the control. All three PD distances are lower than the CNN's on 150 of 168 fields and lower than the control's on 156 of 168. Because the final model and the control share initialization, data, and training schedule, the persistence gain cannot be attributed to continued optimization or data volume.

Mean MT distance also improves, by 3.6\% relative to the CNN and 5.9\% relative to the control. The per-field picture is weaker than for PD: the final model has lower MT distance than the CNN on 98 of 168 fields and lower than the control on 124. We return to this decoupling in Sec.~\ref{sec:complementarity}.

\subsection{Loss study: which losses drive which descriptor}
\label{sec:ablation}
\begin{figure}[t]
\centering
\includegraphics[width=\columnwidth]{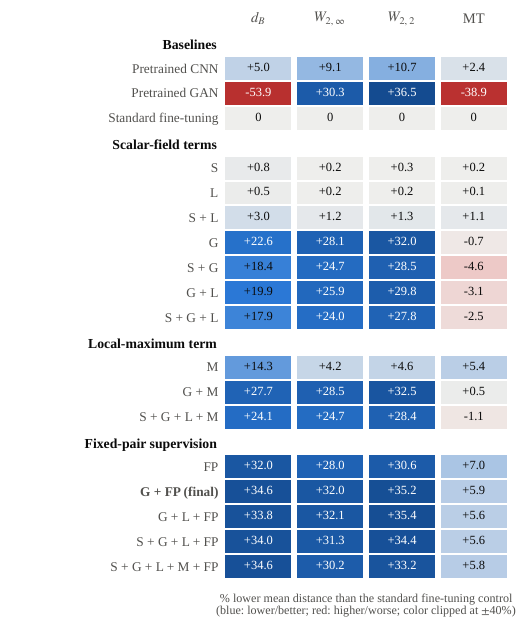}
\caption{Loss study. Each cell gives the percent reduction in a mean distance relative to the standard fine-tuning control (positive and blue: lower distance; red: higher). Configurations are named by the losses added to $L_{uv}$: speed (S), gradient (G), soft level-set (L), local maximum (M), and fixed-pair supervision (FP). Full values are in Appendix~\ref{app:full}.}
\label{fig:heatmap}
\end{figure}

\begin{figure*}[t]
\centering
\includegraphics[width=\textwidth]{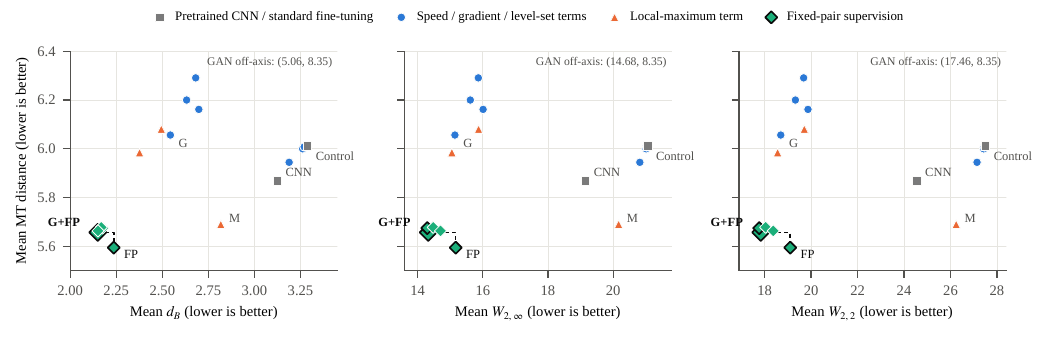}
\caption{Mean PD distance versus mean merge-tree distance for all topology-bearing methods (lower-left is better), for each of the three PD distances. Outlined markers lie on the Pareto front of that pair of distances (dashed). Labels mark the pretrained CNN, the standard fine-tuning control, and the configurations with gradient (G), local-maximum (M), fixed-pair (FP), and gradient plus fixed-pair (G+FP, final) supervision. The pretrained GAN lies outside the plotted range; its coordinates are given in each panel.}
\label{fig:tradeoff}
\end{figure*}

Figure~\ref{fig:heatmap} summarizes all 16 configurations, together with the two pretrained baselines, as the percent change in each mean distance relative to the standard fine-tuning control; full means, ranks, and win counts are in Appendices~\ref{app:full}--\ref{app:wins}. We report every configuration so that the final objective can be judged against the alternatives, including those that are better on individual distances. Figure~\ref{fig:tradeoff} plots each PD distance against MT distance.

To separate the contributions of individual losses we use the factorial structure of the study (Sec.~\ref{sec:study-design}). The \emph{main effect} of a loss is the change in a mean distance when that loss is added, averaged over all combinations of the other losses in the design; an \emph{interaction} measures how much the effect of one loss depends on whether another is present. All effects are listed in Appendix~\ref{app:factorial}.

\textbf{Gradient supervision improves PD agreement but not MT.}
Across the eight combinations of speed, gradient, and level-set losses, adding the gradient loss lowers mean $d_B$, $\Wi$, and $\Wt$ by 0.61, 5.32, and 7.99 on average, while the speed and level-set effects and all interactions are roughly an order of magnitude smaller. Gradient supervision alone lowers $d_B$, $\Wi$, and $\Wt$ by 22.6\%, 28.1\%, and 32.0\% relative to the control. Its average effect on MT, however, is unfavorable (it raises mean MT distance by 0.19): the gradient-only configuration is slightly worse than the control in mean MT (by 0.7\%), and every combination of gradient with speed or level-set losses is worse still (by 2.5\% to 4.6\%). This behavior is consistent with gradient supervision moving the persistence-pair distribution closer to the ground truth without, on its own, preserving how components merge.

\textbf{Critical-location supervision is associated with stronger MT improvement.}
The losses that constrain predicted values at selected critical locations give the largest MT improvements. Adding the local-maximum loss to the control improves MT by 5.4\% and $d_B$ by 14.3\%, but the Wasserstein distances by less than 5\%. Fixed-pair supervision is stronger on the PD distances: added to the control, it lowers $d_B$, $\Wi$, and $\Wt$ by 32.0\%, 28.0\%, and 30.6\% and gives the best mean MT of all 18 methods (5.594, 7.0\% below the control). Small improvements on all four distances also occur for other losses (the local-maximum loss and, marginally, the speed and level-set losses when added to the control), but fixed-pair supervision is the only loss whose main effect is clearly favorable on all four distances in both $2\times2$ designs, and every configuration that contains it improves all four by 5.6\% to 35.4\% relative to the control. It therefore gives the largest and most consistent joint PD and MT improvement in the study. Because $\Lcv$ and $\Lpers$ were always added together, these results describe their combined effect; the study does not separate their individual contributions.

\textbf{Interactions are modest and interpretable.}
Once fixed-pair supervision is present, the local-maximum loss is largely redundant: their interaction is unfavorable on all four distances, and adding the local-maximum loss to S+G+L+FP lowers $d_B$ by only 0.8\% while \emph{raising} both Wasserstein distances by 1.6--1.8\%. This is consistent with fixed-pair supervision already constraining the maxima that $\Lmax$ targets. The level-set loss interacts favorably with fixed-pair supervision on the Wasserstein distances, which is why G+L+FP is marginally better than the final G+FP on $\Wi$ and $\Wt$ (by 0.2--0.4\%) but marginally worse on $d_B$ and MT (by 1.2\% and 0.3\%). Gradient and fixed-pair supervision contribute partly distinct signals. Adding gradient supervision to FP mainly improves the accumulated distances ($\Wi$ by 5.6\%, $\Wt$ by 6.6\%, $d_B$ by 3.9\%) at a small MT cost (1.1\%); adding FP to gradient supervision mainly improves $d_B$ (by 15.5\%) and MT (by 6.6\%).

\subsection{Choosing the final objective}
\label{sec:selection}
No configuration is best on every distance (Appendix~\ref{app:full}): G+FP has the best mean $d_B$, G+L+FP the best $\Wi$ and $\Wt$ among the learned configurations, the pretrained GAN the best $\Wt$ overall, and FP alone the best MT. We use G+FP as the final objective because it gives the best overall balance across the four distances: among all 18 methods it ranks first on $d_B$, second on $\Wi$, third on $\Wt$, and second on MT, whereas G+L+FP ranks 4/1/2/4 and FP alone 5/8/8/1. G+FP lies on the PD--MT Pareto front for every PD distance (Fig.~\ref{fig:tradeoff}), as does FP alone, which remains the better choice when MT is the priority. G+FP is also simpler than the larger fixed-pair configurations: removing the speed and level-set losses from S+G+L+FP slightly improves all four mean distances (by 0.4--1.2\%).

This choice was made after examining the results on the same 168-field benchmark on which they are reported; with no separate validation split, the benchmark served for both evaluation and configuration selection. To limit the resulting optimism, we report all 16 configurations, and the conclusions of Sec.~\ref{sec:headline} hold for each of the five fixed-pair configurations: each lowers all three mean PD distances by 28--35\% relative to the control, improves mean MT by 5.6--7.0\%, and lowers all three PD distances relative to the CNN on 147--153 of 168 fields.

\subsection{Persistence and merge-tree agreement are complementary}
\label{sec:complementarity}
Mean PD and MT improvements do not imply that the two descriptors agree field by field. Of the 150 fields on which the final model lowers all three PD distances relative to the CNN, MT also improves on 91 (60.7\%) and worsens on 59. Relative to the control, MT improves on 119 of the 156 fields with all-three-PD improvement (76.3\%). Some of the largest PD gains coincide with MT losses: on Sample~84, for example, the final model lowers $d_B$ by 55.2\% relative to the CNN while MT distance increases by 8.0\%.

Correlations across methods and fields show the same structure. Across the 18 method means, $d_B$ and MT are strongly correlated (Pearson $r=0.85$), whereas $\Wi$ and $\Wt$ are essentially uncorrelated with MT ($r=-0.13$ and $-0.18$). The distinction is sharper when competing reconstructions of the same field are compared: across methods within individual fields, the median Pearson correlation with MT is 0.70 for $d_B$ but approximately zero for $\Wi$ and $\Wt$ ($-0.02$ and $-0.07$), and after removing additive field and method effects it is 0.73 for $d_B$ versus 0.18 and 0.14. Bottleneck distance and MT thus tend to rank competing reconstructions similarly, whereas the accumulated Wasserstein distances often provide different evidence. One plausible explanation is that $d_B$ and MT are both sensitive to dominant structural discrepancies, whereas $\Wi$ and $\Wt$ accumulate many smaller pairwise mismatches; we offer this as an interpretation rather than an established mechanism. In either case, PD and MT are complementary descriptors rather than interchangeable measures of one ``topological quality.''

\subsection{Fidelity and physics/domain behavior}
Table~\ref{tab:all-metrics} compares the final model with the pretrained CNN and the control on the non-topology measures. Fine-tuning with the reconstruction loss alone improves direct fidelity substantially but worsens spectral, gradient-distribution, and component-structure measures relative to the CNN. The control is best on direct fidelity: PSNR, speed SSIM, speed MAE/RMSE, and WPD MAE. The final model sits between the CNN and the control on these measures. It gains 1.30~dB PSNR over the CNN, about half of the control's 2.60~dB gain, and 0.052 in speed SSIM, about 72\% of the control's gain. The final model therefore retains a substantial share of the fidelity benefit of fine-tuning, but not all of it.

On structure-sensitive measures, the final model is best of the three: it substantially reduces WPD bias and WPD distribution error, gradient MAE and gradient-distribution error, and component-curve error, and it slightly improves PSD log-$L_2$ relative to the CNN. One clear regression remains: PSD slope error is worse than the CNN's (1.139 versus 0.915), improving on only 3 of 168 fields, although it is better than the control's (1.456). Gradient-kurtosis error improves on average but on only 68 of 168 fields.

\begin{table*}[t]
\caption{Non-topology metrics on the 168-field benchmark. Higher is better for PSNR and SSIM; lower is better otherwise. The last three columns give relative changes (positive: better), with PSNR differences in dB: the standard fine-tuning control relative to the CNN, the final model relative to the CNN, and the final model relative to the control.}
\label{tab:all-metrics}
\centering
\scriptsize
\begin{tabular}{lllrrrrrr}
\toprule
Family & Metric & Prefer & CNN & Control & Final & Control vs.\ CNN & Final vs.\ CNN & Final vs.\ control\\
\midrule
Direct & PSNR-$uv$ & $\uparrow$ & 31.1925 & \textbf{33.7892} & 32.4949 & $+2.60$\,dB & $+1.30$\,dB & $-1.29$\,dB\\
Direct & SSIM-speed & $\uparrow$ & 0.7412 & \textbf{0.8134} & 0.7932 & $+9.8\%$ & $+7.0\%$ & $-2.5\%$\\
Direct & Speed MAE & $\downarrow$ & 0.6941 & \textbf{0.4958} & 0.5899 & $+28.6\%$ & $+15.0\%$ & $-19.0\%$\\
Direct & Speed RMSE & $\downarrow$ & 1.1078 & \textbf{0.8555} & 0.9891 & $+22.8\%$ & $+10.7\%$ & $-15.6\%$\\
\midrule
Wind power & WPD bias $|\cdot|$ & $\downarrow$ & 35.3439 & 31.3033 & \textbf{22.6377} & $+11.4\%$ & $+36.0\%$ & $+27.7\%$\\
Wind power & WPD MAE & $\downarrow$ & 231.6709 & \textbf{158.9769} & 190.6897 & $+31.4\%$ & $+17.7\%$ & $-19.9\%$\\
Wind power & WPD W1 & $\downarrow$ & 45.2713 & 39.3927 & \textbf{24.7322} & $+13.0\%$ & $+45.4\%$ & $+37.2\%$\\
\midrule
Spectrum & PSD log-$L_2$ & $\downarrow$ & 0.8335 & 1.1338 & \textbf{0.8147} & $-36.0\%$ & $+2.3\%$ & $+28.1\%$\\
Spectrum & PSD slope $|\Delta|$ & $\downarrow$ & \textbf{0.9150} & 1.4560 & 1.1385 & $-59.1\%$ & $-24.4\%$ & $+21.8\%$\\
\midrule
Gradients & Gradient MAE & $\downarrow$ & 0.3491 & 0.3247 & \textbf{0.3125} & $+7.0\%$ & $+10.5\%$ & $+3.8\%$\\
Gradients & Gradient W1 & $\downarrow$ & 0.2329 & 0.2496 & \textbf{0.1381} & $-7.2\%$ & $+40.7\%$ & $+44.7\%$\\
Gradients & Gradient kurtosis $|\Delta|$ & $\downarrow$ & 3.7004 & 5.5793 & \textbf{3.2601} & $-50.8\%$ & $+11.9\%$ & $+41.6\%$\\
\midrule
Structure & Component curve $L_1$ & $\downarrow$ & 115.5278 & 141.9157 & \textbf{81.1815} & $-22.8\%$ & $+29.7\%$ & $+42.8\%$\\
\bottomrule
\end{tabular}
\end{table*}

\subsection{Qualitative interpretation of persistence improvements}
\label{sec:qualitative-pd}
To reduce retrospective example-selection bias, we first ranked the 168 fields using conventional fidelity alone. For each field, we converted the absolute PSNR difference and the relative speed-MAE and speed-RMSE differences between the final model and the pretrained CNN to percentile ranks across fields, and defined conventional closeness as the maximum of the three ranks, so that a field counts as close only if it is close under all three measures. The 34 closest fields (20\% of the benchmark) form the primary near-tie cohort; no topology measure entered this selection. Only after freezing the cohort did we examine topology, requiring simultaneous improvement in $d_B$, $\Wi$, and $\Wt$ and ranking eligible fields by the smallest of the three relative improvements. All 34 fields in the cohort improve on the pretrained CNN under all three PD distances, as do all fields in a stricter 17-field and a broader 51-field cohort. We show five predeclared fields from this cohort (Samples 78, 71, 80, 63, and 69), selected before image inspection to span strong topology consensus, very high conventional closeness, large bottleneck gains, and different positions in the near-tie ranking. Relative to the control, the final model lowers $d_B$ by 28.7--50.2\%, $\Wi$ by 31.5--37.2\%, and $\Wt$ by 34.0--39.7\% on all five fields, while MT gains range from 0.6\% (Sample~69) to 15.7\% (Sample~71).

\begin{figure*}[t]
    \centering
    \includegraphics[width=\textwidth]{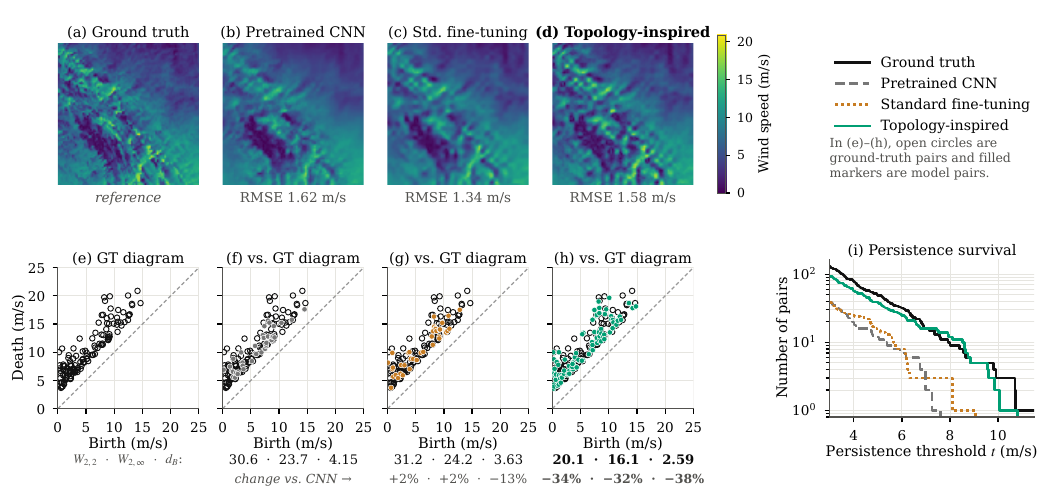}
    \caption{Qualitative comparison for Sample~78, a field on which the final model and the pretrained CNN are nearly tied in conventional error. (a--d)~Wind speed on the $160\times160$ topology crop for the ground truth, pretrained CNN, standard fine-tuning control, and topology-inspired final model (G+FP), on a shared color scale; RMSE is computed on the crop. (e--h)~Persistence diagrams (finite $D_0$ and $D_1$ pairs combined): the ground-truth pairs (open circles) alone and overlaid with each model's pairs (filled). Only pairs with persistence of at least 3~m/s are drawn, but the distances below each panel ($\Wt$, $\Wi$, $d_B$) use the complete finite diagrams; percentages are changes relative to the pretrained CNN. (i)~Number of pairs whose persistence is at least $t$.}
    \label{fig:qualitative-sample78}
\end{figure*}

Figure~\ref{fig:qualitative-sample78} shows Sample~78, the principal example. Its RMSE is 1.58 for the final model and 1.62 for the pretrained CNN, while the control attains the lowest pointwise error (RMSE 1.34). Despite near-identical conventional error, the final model lowers $\Wt$, $\Wi$, and $d_B$ relative to the CNN by 34\%, 32\%, and 38\%, whereas the control changes them by $+2\%$, $+2\%$, and $-13\%$. The persistence-diagram overlays show why: the final model's diagram extends into the long-lived region occupied by the ground truth, while the CNN and control diagrams collapse toward the diagonal. The survival curve, which counts pairs whose persistence exceeds a threshold, confirms that the final model better matches the ground truth's long-lived tail. A reconstruction is not preferable merely because it contains more pairs, since additional pairs can also represent unsupported fragmentation; the distances above, computed on the complete diagrams, measure agreement in birth--death coordinates rather than feature count.

\section{Discussion}

\subsection{What the objective is measuring}
We read the loss study as separating two kinds of signal; what follows is an interpretation consistent with the results rather than an established mechanism. Gradient supervision changes the distribution of sharp transitions in the predicted field, counteracting the smoothing introduced by standard fine-tuning. This is consistent with restoring moderately and highly persistent features whose birth--death coordinates strongly influence the accumulated Wasserstein distances. Gradient supervision does not, however, constrain where critical values lie or what their values are, and its unfavorable MT effect suggests that sharper fields can merge components in a different order. Fixed-pair supervision does constrain values, at both endpoints of every supervised ground-truth pair, including saddle-side endpoints, and it constrains the persistence of those pairs. This plausibly explains why it improves worst-case PD mismatch and MT together.

Fixed-pair supervision still leaves important structure unsupervised. It fixes which ground-truth vertices are constrained but not which features the prediction actually forms there. Unlike re-matched diagram losses~\cite{Kissi2025}, it cannot penalize spurious pairs that the prediction creates elsewhere, and like all of our losses it does not constrain ancestry or branch assignment. These gaps may contribute to residual PD disagreement, and they do not address the hierarchy measured by MT; MT worsens on 59 of the 150 fields whose PD distances all improve relative to the CNN. The fixed-pair formulation includes supervision at saddle-side endpoints and improves MT on average, but because $\Lcv$ and $\Lpers$ were always combined, the source of this improvement is not isolated; hierarchy itself remains outside the objective.

The visually small field-space differences between the fine-tuned models do not imply that the PD result is meaningless. Persistence coordinates can change through localized shifts in critical values even when global pixelwise differences remain small. At the same time, a denser diagram or a lower PD score alone would not establish scientific improvement. The matched control, agreement across three PD distances with different emphases, coordinate-level diagram overlays, the independent validation of the PD computation, and concurrent gains in several domain-structure metrics together provide evidence that the auxiliary objective is not merely increasing feature count or exploiting a single evaluation measure.

\subsection{Limitations}
\textbf{Scope.}
The study uses one pretrained SR framework, one spatial region, and one 168-hour benchmark. The fine-tuning set improves seasonal coverage but does not establish geographic or architectural generalization. All configurations are compared at a single fine-tuning scale of 2688 samples. Earlier development runs at smaller scales used a different topology-inspired objective and the PD-distance convention of the workshop submission, so they are excluded from this analysis; those trained models remain valid and could be re-evaluated separately.

\textbf{Selection and variance.}
The final objective was chosen on the evaluation benchmark itself (Sec.~\ref{sec:selection}); a held-out validation split would be preferable. Each configuration was trained once, so run-to-run variance is not quantified, and apart from the reduction of the fixed-pair weights described in Sec.~\ref{sec:study-design}, no weight search was performed.

\textbf{Supervision.}
$\Lcv$ and $\Lpers$ were always used together, so the study does not isolate their individual contributions. Fixed-pair supervision operates only at ground-truth-selected endpoint locations and does not penalize topological features that arise elsewhere in the reconstruction; a re-matched diagram loss could complement it by identifying and suppressing such unsupported features. The final model regresses relative to the pretrained CNN on PSD slope and retains only part of the control's fidelity gain.

\textbf{Topology crop.}
PD and MT are computed on a fixed $160\times160$ subregion of each $500\times500$ reconstruction. This keeps the topology pipeline tractable and all methods directly comparable, but the topological conclusions apply to that region and do not establish preservation over the full domain; the full field or multiple crops should be evaluated in future work.

\textbf{Descriptors.}
PD and MT are computed on scalar speed rather than on the two-channel training target. Our independent validation covers the PD distances and merge-tree construction, but not the merge-tree distance computation itself. Most importantly for applications, persistence diagrams deliberately discard spatial embedding. Two features with similar birth and death values can match closely in a PD even when they occur at very different physical locations: a pair born at 4~m/s and dying at 12~m/s near one corner of the domain matches an identical pair near the opposite corner at zero cost. This invariance is useful for topological comparison but is a limitation for wind-field applications, in which the geographic location of a reconstructed structure may be scientifically consequential. Neither PD nor MT distance localizes where structure is preserved or displaced, and downstream scientific tasks are needed to establish the practical meaning of the gains.

\subsection{Future directions}
\textbf{Generalization.}
The findings should be tested with larger evaluation benchmarks and fine-tuning sets spanning more time periods and geographically distinct regions, with repeated runs, and with other SR architectures. Generative models such as the pretrained PhIRE GAN are a natural target: they already match the accumulated Wasserstein distances well, but their bottleneck and MT agreement is weak. Beyond fine-tuning, the most effective losses could be used to train SR models from the start.

\textbf{Topology-aware learning.}
Hierarchy-aware supervision could target what the present losses leave out. Options range from fixed ground-truth constraints on branch and merge relationships, the hierarchy analogue of fixed-pair supervision, to merge-tree losses in which the prediction's tree is recomputed and matched at each step, analogous to re-matched diagram losses. Learned MT similarity models such as MTNN~\cite{QinMTNN2024} could also provide a surrogate hierarchy objective, provided that the extracted tree representation is connected back to differentiable scalar-field values during optimization and that the model's accuracy is validated for wind-speed fields. Re-matched diagram losses~\cite{Kissi2025} could complement fixed-pair supervision by suppressing unsupported features. Separating $\Lcv$ from $\Lpers$ and studying the loss weights systematically would clarify which parts of fixed-pair supervision matter. Controlled synthetic fields with analytically known critical points and merge structure (for example, two peaks merging, a saddle moving, a weak feature appearing, or identical diagrams with different merge hierarchies) would provide a complementary test of which losses recover which topological events, independently of the complexity of atmospheric data. Finally, extending supervision and evaluation from scalar wind speed to vector-field topology would connect this work to vector-field SR more directly.

\textbf{Spatial and scientific validation.}
Mapping matched persistence pairs back to their physical locations would allow persistent-feature preservation to be measured as displacement, splitting, merging, appearance, and disappearance of specific structures, rather than as a single scalar distance. It would also connect topological agreement to regions of practical interest, such as wind-farm sites and meteorological stations whose observations could provide physical validation. We are pursuing spatially aware persistence evaluation as a separate study.

\section{Conclusion}
We presented a controlled proof of concept for topology-inspired fine-tuning in wind-field SR. Fine-tuning with the reconstruction loss alone worsens all three mean PD distances; adding gradient-magnitude and fixed-pair supervision reverses this, lowering them by 25--31\% relative to the pretrained CNN and 32--35\% relative to the matched control while retaining about half of the control's PSNR gain. Merge-tree distance improves on average but much less consistently, and different PD distances can prefer different reconstructions, so topological quality is descriptor- and distance-dependent. Within this single-model, single-region setting, the loss study shows that gradient supervision mainly improves persistence agreement whereas fixed-pair supervision gives the most consistent joint PD and MT improvement, pointing toward hierarchy-aware and spatially aware supervision and evaluation as the next steps.

\raggedbottom
\section*{Acknowledgments}
Wind-field data were sourced from the NREL WIND Toolkit (NREL, now the National Laboratory of the Rockies, NLR). Topological descriptors were computed using the Topology ToolKit.

\bibliographystyle{abbrv-doi}
\bibliography{template}

\clearpage
\onecolumn
\appendix

\section{Full loss-study results}
\label{app:full}
\begin{table}[H]
\caption{All topology-bearing methods on the 168-field benchmark (means; lower is better). Configurations are named by the losses added to the reconstruction loss $L_{uv}$; loss weights are fixed across configurations (Sec.~\ref{sec:study-design}). The middle columns give the percent reduction in mean distance relative to the standard fine-tuning control (blue: lower distance; red: higher). Best value per column in bold, second best underlined. Ranks are over all 18 methods for $d_B$/$\Wi$/$\Wt$/MT.}
\label{tab:all-candidates}
\centering
\scriptsize
\setlength{\tabcolsep}{4pt}
\begin{tabular}{lrrrrrrrrc}
\toprule
 & \multicolumn{4}{c}{Mean distance} & \multicolumn{4}{c}{\% reduction vs.\ control} & \\
\cmidrule(lr){2-5}\cmidrule(lr){6-9}
Method & $d_B$ & $\Wi$ & $\Wt$ & MT & $d_B$ & $\Wi$ & $\Wt$ & MT & Ranks\\
\midrule
\input{tables/tab_all_candidates.tex} 
\bottomrule
\end{tabular}
\end{table}

\section{Loss-weight calibration}
\label{app:weights}
\begin{table}[H]
\caption{Loss-scale calibration at the pretrained CNN checkpoint before fine-tuning. Means are computed over all 168 fields of the fixed benchmark in 42 batches of four, evaluating the losses without updating model weights. Shown are the mean unweighted loss, its coefficient, and the weighted contribution relative to $L_{uv}$. Coefficients compensate for the very different numerical scales of the unweighted losses.}
\centering
\scriptsize
\begin{tabular}{lrrrr}
\toprule
Loss & Unweighted mean & Coefficient & Weighted & Relative to $L_{uv}$\\
\midrule
$L_{uv}$ & 0.031759 & 1 & 0.031759 & 100\%\\
$L_{\mathrm{speed}}$ & 1.270521 & 0.01 & 0.012705 & 40\%\\
$L_{\mathrm{grad}}$ & 0.575292 & 0.05 & 0.028765 & 91\%\\
$L_{\mathrm{levelset}}$ & 0.034349 & 0.25 & 0.008587 & 27\%\\
$\Lmax$ & 6.064671 & 0.001 & 0.006065 & 19\%\\
\bottomrule
\end{tabular}
\end{table}
Fixed-pair supervision was introduced after this diagnostic and is therefore not included in the table. During training of the final G+FP model, representative weighted contributions $(0.004\,\Lcv,\;0.002\,\Lpers)$ were $(0.0103,\,0.0061)$ at iteration 2006 and $(0.0259,\,0.0186)$ at iteration 2016, comparable in scale to $L_{uv}$ ($0.0106$ and $0.0214$) and $0.05\,L_{\mathrm{grad}}$ ($0.0092$ and $0.0184$) at the same iterations.

\section{Factorial effects}
\label{app:factorial}
Each effect is the average change in a mean distance attributable to a loss (main effect) or to the combination of two or three losses beyond their separate effects (interaction), estimated from all configurations in the design with standard $\pm1$ coding. Effects are reported for the negated distance, so a positive value means the loss, or the interaction, lowers that distance on average; units are those of each distance.
\begin{table}[H]
\caption{Factorial effects on the four topology distances (positive: lowers the distance on average). (a)~All eight combinations of the speed, gradient, and level-set losses. (b)~The local-maximum loss crossed with fixed-pair supervision, both added to speed + gradient + level-set. (c)~The level-set loss crossed with fixed-pair supervision, both added to gradient supervision.}
\label{tab:factorial}
\centering
\scriptsize
\setlength{\tabcolsep}{4pt}
\begin{tabular}{lrrrr}
\toprule
Effect & $d_B$ & $\Wi$ & $\Wt$ & MT\\
\midrule
\input{tables/tab_factorial.tex} 
\bottomrule
\end{tabular}
\end{table}

\section{Median distances}
\label{app:medians}
\begin{table}[H]
\caption{Median distances across the 168 fields (lower is better). The main conclusions agree with the means in Table~\ref{tab:all-candidates}, although some close rankings change (for example, G+FP has a lower median $\Wi$ than G+L+FP): G+FP has the lowest median $d_B$, FP alone the lowest median MT, and every fixed-pair configuration improves on the control and the CNN in all four medians.}
\centering
\scriptsize
\begin{tabular}{lrrrr}
\toprule
Method & $d_B$ & $\Wi$ & $\Wt$ & MT\\
\midrule
\input{tables/tab_medians.tex} 
\bottomrule
\end{tabular}
\end{table}

\section{Per-field win counts}
\label{app:wins}
\begin{table}[H]
\caption{Number of the 168 fields on which each method has a lower distance than the pretrained CNN (left) and than the standard fine-tuning control (right). ``3PD'': all three PD distances lower; ``+MT'': all three PD distances and MT lower.}
\centering
\scriptsize
\setlength{\tabcolsep}{4pt}
\begin{tabular}{lrrrrrrrrrrrr}
\toprule
 & \multicolumn{6}{c}{vs.\ pretrained CNN} & \multicolumn{6}{c}{vs.\ control}\\
\cmidrule(lr){2-7}\cmidrule(lr){8-13}
Method & $d_B$ & $\Wi$ & $\Wt$ & MT & 3PD & +MT & $d_B$ & $\Wi$ & $\Wt$ & MT & 3PD & +MT\\
\midrule
\input{tables/tab_wins.tex} 
\bottomrule
\end{tabular}
\end{table}

\section{Targeted contrasts}
\label{app:contrasts}
\begin{table}[H]
\caption{Percent reduction in mean distance from adding or removing losses (positive: the change lowers the distance). S: speed, G: gradient, L: level-set, M: local maximum, FP: fixed-pair supervision. These are descriptive contrasts between trained models, not causal effect estimates.}
\centering
\scriptsize
\setlength{\tabcolsep}{3pt}
\begin{tabular}{lrrrr}
\toprule
Change & $d_B$ & $\Wi$ & $\Wt$ & MT\\
\midrule
\input{tables/tab_contrasts.tex} 
\bottomrule
\end{tabular}
\end{table}

\end{document}

%% file: figures/topology_descriptor_schematic_tikz.tex
\begin{figure*}[t]
\centering
\resizebox{0.98\textwidth}{!}{%
\begin{tikzpicture}[
  >=Latex,
  panel/.style={draw=black!45, rounded corners=2pt, fill=white, minimum width=4.95cm, minimum height=4.85cm},
  title/.style={font=\bfseries\scriptsize, align=center},
  small/.style={font=\scriptsize},
  tinytext/.style={font=\tiny, align=center},
  minpt/.style={circle, fill=blue!65, draw=white, line width=0.4pt, minimum size=0.20cm, inner sep=0pt},
  maxpt/.style={circle, fill=red!75, draw=white, line width=0.4pt, minimum size=0.20cm, inner sep=0pt},
  pdpt/.style={circle, fill=pink!90!black, draw=black!15, line width=0.4pt, minimum size=0.22cm, inner sep=0pt},
  treemin/.style={circle, draw=black!55, fill=blue!25, minimum size=0.48cm, inner sep=1pt, font=\scriptsize},
  treemax/.style={circle, draw=black!55, fill=red!28, minimum size=0.48cm, inner sep=1pt, font=\scriptsize},
  treeevent/.style={circle, draw=black!55, fill=orange!30, minimum size=0.48cm, inner sep=1pt, font=\scriptsize},
  edge/.style={thick, black!55}
]

\node[panel] (pA) at (0,0) {};
\node[panel] (pB) at (6.05,0) {};
\node[panel] (pC) at (12.10,0) {};

\node[title] at ($(pA.north)+(0,-0.32)$) {(a) Scalar field and threshold sweep};
\begin{scope}[shift={(-2.12,-1.55)}, x=0.59cm, y=0.59cm]
  \draw[->] (0,0) -- (6.85,0) node[right,small] {$x$};
  \draw[->] (0,0) -- (0,4.75) node[above,small] {$s(x)$};
  \foreach \y in {0.85,1.55,2.55,3.55}
    \draw[dashed,gray!45] (0,\y) -- (6.65,\y);
  \draw[thick,orange!90!black]
    plot[smooth] coordinates {(0.30,3.60) (1.05,1.02) (1.70,3.02) (2.75,0.52) (3.75,2.52) (4.90,1.42) (6.35,4.05)};
  \node[minpt,label={[small]below:$m_1$}] at (1.05,1.02) {};
  \node[maxpt,label={[small]above:$M_1$}] at (1.70,3.02) {};
  \node[minpt] at (2.75,0.52) {};
  \node[small, below=3pt, fill=white, inner sep=1pt] at (2.75,0.52) {$m_2$};
  \node[maxpt,label={[small]above:$M_2$}] at (3.75,2.52) {};
  \node[minpt,label={[small]below:$m_3$}] at (4.90,1.42) {};
  \node[maxpt,label={[small]above:$M_3$}] at (6.35,4.05) {};
\end{scope}

\node[title] at ($(pB.north)+(0,-0.32)$) {(b) Persistence diagram};
\begin{scope}[shift={(4.15,-1.80)}, x=0.60cm, y=0.60cm]
  \draw[->] (0,0) -- (6.25,0) node[right,small] {birth};
  \draw[->] (0,0) -- (0,5.30);
  \node[small,rotate=90,anchor=center] at (-0.70,2.55) {death};
  \draw[dashed,gray!70] (0,0) -- (4.70,4.70);
  \node[gray,font=\tiny,anchor=north west] at (3.85,4.00) {$d{=}b$};
  \node[pdpt] (q1) at (1.00,1.90) {};
  \node[pdpt] (q2) at (2.50,3.85) {};
  \node[pdpt] (q3) at (3.80,4.65) {};
  \node[small,anchor=west] at (1.22,1.87) {$p_1$};
  \node[small,anchor=west,fill=white,inner sep=1pt] at (2.85,3.65) {$p_2{=}(b_2,d_2)$};
  \node[small,anchor=west,fill=white,inner sep=1pt] at (4.10,4.72) {$p_3$};
  \draw[<->,thick] (2.18,2.50) -- (2.18,3.85);
  \node[small,anchor=east] at (1.96,3.17) {$d_2{-}b_2$};
  \draw[dotted,gray!60] (q2) -- (2.50,2.50);
\end{scope}

\node[title] at ($(pC.north)+(0,-0.32)$) {(c) Merge-tree hierarchy};
\begin{scope}[shift={(10.00,-1.45)}]
  \node[treemax]   (root) at (2.05,3.10) {$M_3$};
  \node[treeevent] (e1)   at (2.05,2.22) {$M_1$};
  \node[treemin]   (m1)   at (0.95,0.72) {$m_1$};
  \node[treeevent] (e2)   at (3.05,1.55) {$M_2$};
  \node[treemin]   (m2)   at (2.55,0.25) {$m_2$};
  \node[treemin]   (m3)   at (3.78,0.58) {$m_3$};
  \draw[edge] (root) -- (e1);
  \draw[edge] (e1) -- (m1);
  \draw[edge] (e1) -- (e2);
  \draw[edge] (e2) -- (m2);
  \draw[edge] (e2) -- (m3);
  \draw[<->,gray!55,thin] (1.52,2.26) -- (1.52,3.06);
  \node[font=\tiny,gray,anchor=east,align=right] at (1.44,2.66) {scalar\\interval};
\end{scope}

\node[tinytext] at (6.05,-2.80)
  {\textcolor{orange!90!black}{\rule[0.55ex]{1em}{0.8pt}} scalar field \quad
   \textcolor{blue!65}{$\bullet$} minima\,/\,births \quad
   \textcolor{red!75}{$\bullet$} maxima\,/\,root \quad
   \textcolor{orange!70!black}{$\bullet$} merge events \quad
   \textcolor{pink!90!black}{$\bullet$} PD pairs};

\draw[->,thick,black!55] ($(pA.east)+(0.05,0)$) -- ($(pB.west)+(-0.05,0)$);
\draw[->,thick,black!55] ($(pB.east)+(0.05,0)$) -- ($(pC.west)+(-0.05,0)$);

\end{tikzpicture}%
}
\caption{Conceptual relationship among scalar threshold sweeps, persistence diagrams, and merge trees. The scalar field produces critical-point events as the threshold changes. A persistence diagram keeps paired birth/death values and persistence, whereas a merge tree additionally preserves ancestry and merge order.}
\label{fig:topology-schematic}
\end{figure*}

%% file: figures/data_pipeline_schematic_tikz.tex
\begin{figure*}[t]
\centering
\resizebox{0.98\textwidth}{!}{%
\begin{tikzpicture}[
  >=Latex,
  box/.style={draw=black!45, rounded corners=3pt, fill=white, align=center,
              font=\scriptsize, minimum width=3.10cm, minimum height=1.10cm, text width=2.85cm},
  artifact/.style={draw=black!45, rounded corners=3pt, fill=blue!5, align=center,
                   font=\scriptsize, minimum width=3.10cm, minimum height=1.10cm, text width=2.85cm},
  ctrl/.style={draw=black!50, rounded corners=3pt, fill=orange!10, align=center,
               font=\scriptsize, minimum width=3.60cm, minimum height=1.00cm, text width=3.30cm},
  topo/.style={draw=teal!55, line width=0.8pt, rounded corners=3pt, fill=teal!7, align=center,
               font=\scriptsize, minimum width=3.60cm, minimum height=1.00cm, text width=3.50cm},
  eval/.style={draw=black!45, rounded corners=3pt, fill=green!8, align=center,
               font=\scriptsize, minimum width=3.10cm, minimum height=1.10cm, text width=2.85cm},
  head/.style={font=\bfseries\scriptsize, align=center},
  arrow/.style={->, very thick, black!65},
  sarrow/.style={->, thick, black!50}
]

\node[box]      (n1) at ( 0.00, 1.80) {\textbf{1.~WIND Toolkit}\\HSDS stream\\100\,m wind fields};
\node[box]      (n2) at ( 3.65, 1.80) {\textbf{2.~Vector conversion}\\speed + direction\\$\rightarrow[u,v]$};
\node[artifact] (n3) at ( 7.30, 1.80) {\textbf{3.~PhIRE hierarchy}\\LR$\rightarrow$MR, MR$\rightarrow$HR\\paired data};
\node[artifact] (n4) at (10.95, 1.80) {\textbf{4.~MR--HR pairs}\\paired fields\\$100\!\times\!100\rightarrow 500\!\times\!500$};
\draw[arrow] (n1) -- (n2);
\draw[arrow] (n2) -- (n3);
\draw[arrow] (n3) -- (n4);

\node[artifact] (n5) at ( 0.00,-1.20) {\textbf{5.~Data split}\\168-field evaluation benchmark\\2688-sample seasonal fine-tuning set};
\node[ctrl]     (n6a)at ( 4.40,-0.45) {\textbf{6a.~Standard}\\\textbf{fine-tuning control}\\$L_{uv}$ only};
\node[topo]     (n6b)at ( 4.40,-2.00) {\textbf{6b.~Topology-inspired}\\$L_{uv}{+}L_{\text{grad}}{+}L_{\text{CV}}{+}L_{\text{pers}}$\\[-1pt]{\tiny + 14 further loss configurations}};
\node[eval]     (n7) at ( 8.75,-1.20) {\textbf{7.~Inference}\\all models evaluated on\\same 168 fields};
\node[eval]     (n8) at (12.40,-1.20) {\textbf{8.~Metrics}\\fidelity / domain (full field)\\PD ($d_B$, $W_{2,\infty}$, $W_{2,2}$) \& MT\\on $160\!\times\!160$ crop};

\draw[arrow] (n4.south) -- ++(0,-0.50) -- ++(-10.95,0) -- (n5.north);

\node[head] at (5.47, 2.95) {Data construction};
\node[head] at (6.38, 0.30) {Training/evaluation and measurement};

\coordinate (jct) at (2.10,-1.20);
\draw[sarrow] (n5.east) -- (jct);
\draw[sarrow] (jct) |- (n6a.west);
\draw[sarrow] (jct) |- (n6b.west);

\coordinate (jct2) at (6.90,-1.20);
\draw[sarrow] (n6a.east) -| (jct2);
\draw[sarrow] (n6b.east) -| (jct2);
\draw[sarrow] (jct2) -- (n7.west);

\draw[arrow] (n7) -- (n8);

\node[font=\tiny, gray, align=center] at (4.40,-1.22)
  {same pretrained CNN init,\\optimizer \& schedule};

\end{tikzpicture}%
}
\caption{Overview of the data-generation, fine-tuning, and evaluation pipeline. Wind Toolkit fields are streamed through HSDS, converted to two-channel vector fields, and organized into PhIRE-compatible paired-resolution fields (MR$\rightarrow$HR). Every fine-tuned model (the standard fine-tuning control, the final topology-inspired objective, and the other loss configurations of the loss study) shares the same pretrained CNN initialization, 2688-sample fine-tuning set, optimizer, and schedule, and differs only in its auxiliary loss terms. All models, including the pretrained CNN and GAN, are evaluated on the same 168-field benchmark: fidelity and domain measures on the full fields, and persistence-diagram and merge-tree measures on a fixed $160\times160$ crop.}
\label{fig:data-pipeline}
\end{figure*}

%% file: figures/fixed_pair_objective_tikz.tex
\begin{figure*}[t]
\centering
\resizebox{0.98\textwidth}{!}{%
\begin{tikzpicture}[
  >=Latex,
  off/.style={draw=black!45, rounded corners=3pt, fill=blue!5, align=center, font=\scriptsize,
              minimum height=1.15cm, text width=3.05cm},
  fix/.style={draw=black!45, rounded corners=3pt, fill=orange!12, align=center, font=\scriptsize,
              minimum height=1.15cm, text width=3.05cm},
  trn/.style={draw=teal!60, line width=0.7pt, rounded corners=3pt, fill=teal!7, align=center, font=\scriptsize,
              minimum height=1.15cm, text width=3.05cm},
  loss/.style={draw=black!45, rounded corners=3pt, fill=green!8, align=center, font=\scriptsize,
               minimum height=1.15cm, text width=4.75cm},
  interp/.style={draw=black!30, rounded corners=3pt, fill=yellow!10, align=center, font=\scriptsize,
                 minimum height=1.00cm, text width=5.10cm},
  head/.style={font=\bfseries\scriptsize, align=center},
  tinytext/.style={font=\tiny, align=center},
  arrow/.style={->, very thick, black!65},
  sarrow/.style={->, thick, black!55},
  bpt/.style={circle, fill=blue!65, draw=white, line width=0.4pt, minimum size=0.20cm, inner sep=0pt},
  dpt/.style={circle, fill=orange!85!black, draw=white, line width=0.4pt, minimum size=0.20cm, inner sep=0pt}
]

\node[head] at (2.05, 1.15) {Offline, once per ground-truth training field (TTK)};
\node[head] at (11.95, 1.15) {Every training step (differentiable)};
\draw[black!25, dashed] (7.85, 1.35) -- (7.85, -6.15);

\node[off]  (gt)   at ( 0.00, 0) {Ground-truth speed field\\$s_{GT}$ (denormalized)};
\node[off]  (pd)   at ( 3.85, 0) {TTK persistence diagram\\of $s_{GT}$: pairs $(b_i,d_i)$};
\node[fix]  (idx)  at ( 3.85,-1.75) {Fixed endpoint vertices\\$(v^b_i,\,v^d_i)$ and GT values\\$b_i,\ d_i,\ p_i=|d_i-b_i|$};
\node[trn]  (sr)   at ( 9.95, 0) {SR prediction\\$\hat s=\lVert[\hat u,\hat v]\rVert$};
\node[trn]  (gat)  at ( 9.95,-1.75) {Gather at fixed vertices\\$\hat b_i=\hat s(v^b_i)$,\ $\hat d_i=\hat s(v^d_i)$};
\node[loss] (lcv)  at (14.55, 0) {$L_{\mathrm{CV}}=\tfrac12\bigl[\mathrm{MSE}(\hat b,b)+\mathrm{MSE}(\hat d,d)\bigr]$};
\node[loss] (lp)   at (14.55,-1.75) {$L_{\mathrm{pers}}=\mathrm{MSE}(\hat p,p)$,\\ $\hat p_i=|\hat d_i-\hat b_i|$};

\draw[arrow] (gt) -- (pd);
\draw[arrow] (pd) -- (idx);
\draw[arrow] (sr) -- (gat);
\draw[arrow] (idx.east) -- node[above,tinytext]{indices stored,} node[below,tinytext]{never re-matched} (gat.west);
\draw[arrow] (gat.east) -- ++(0.20,0) |- (lcv.west);
\draw[arrow] (gat.east) -- (lp.west);

\begin{scope}[shift={(-1.30,-6.05)}, x=0.78cm, y=0.62cm]
  \draw[->,black!60] (0,0) -- (9.2,0) node[right,font=\scriptsize] {$x$};
  \draw[->,black!60] (0,0) -- (0,5.0) node[above,font=\scriptsize] {$s$};
  \draw[thick,black] plot[smooth] coordinates {(0.2,3.4) (1.4,1.0) (2.6,2.2) (3.6,3.6) (4.8,1.9) (6.0,0.7) (7.4,3.0) (8.8,4.3)};
  \draw[thick,teal!80!black,dashed] plot[smooth] coordinates {(0.2,3.2) (1.4,1.6) (2.6,2.1) (3.6,3.0) (4.8,2.0) (6.0,1.3) (7.4,2.8) (8.8,4.1)};
  \draw[gray!60,densely dotted] (6.0,0) -- (6.0,4.6);
  \draw[gray!60,densely dotted] (3.6,0) -- (3.6,4.6);
  \node[font=\tiny,gray,below] at (6.0,0) {$v^b$ (fixed)};
  \node[font=\tiny,gray,below] at (3.6,0) {$v^d$ (fixed)};
  \node[bpt] at (6.0,0.7) {};
  \node[dpt] at (3.6,3.6) {};
  \node[circle,draw=teal!80!black,fill=white,minimum size=0.17cm,inner sep=0pt] at (6.0,1.3) {};
  \node[circle,draw=teal!80!black,fill=white,minimum size=0.17cm,inner sep=0pt] at (3.6,3.0) {};
  \node[font=\scriptsize,anchor=east] at (5.92,0.55) {$b$};
  \node[font=\scriptsize,anchor=west] at (6.40,1.15) {$\hat b$};
  \node[font=\scriptsize,anchor=south] at (3.6,3.70) {$d$};
  \node[font=\scriptsize,anchor=north] at (3.45,2.85) {$\hat d$};
  \draw[<->,red!70!black,thick] (6.25,0.7) -- (6.25,1.3);
  \draw[<->,red!70!black,thick] (3.85,3.0) -- (3.85,3.6);
  \draw[|-|,black,thick] (9.6,0.7) -- (9.6,3.6) node[midway,right,font=\scriptsize] {$p$};
  \draw[|-|,teal!80!black,thick] (10.6,1.3) -- (10.6,3.0) node[midway,right,font=\scriptsize] {$\hat p$};
  \draw[gray!50,densely dotted] (6.0,0.7) -- (9.6,0.7);
  \draw[gray!50,densely dotted] (3.6,3.6) -- (9.6,3.6);
  \draw[thick,black] (0.4,-1.15) -- (1.0,-1.15) node[right,font=\tiny] {GT $s$};
  \draw[thick,teal!80!black,dashed] (2.1,-1.15) -- (2.7,-1.15) node[right,font=\tiny] {SR $\hat s$};
  \draw[<->,red!70!black,thick] (3.9,-1.15) -- (4.5,-1.15) node[right,font=\tiny] {penalized by $L_{\mathrm{CV}}$};
\end{scope}

\node[interp] (i1) at (12.15,-3.70)
  {Endpoints are critical points of $s_{GT}$ (e.g., an extremum and the saddle at which its feature dies), so saddle-side \emph{values} are supervised as well as extrema.};
\node[interp] (i2) at (12.15,-5.55)
  {Not a PD-distance loss: no diagram of $\hat s$ is computed and nothing is re-matched, so spurious pairs elsewhere are not penalized; PD and MT distances are evaluation measures only.};
\draw[sarrow] (lp.south) -- (i1.north -| lp.south);
\draw[sarrow] (i1) -- (i2);

\end{tikzpicture}%
}
\caption{Fixed-pair supervision used in the final objective. For each ground-truth training field, TTK computes the persistence diagram once, offline, and stores the two endpoint vertices of each supervised persistence pair together with the ground-truth birth value $b$, death value $d$, and persistence $p$. During training, the predicted speed $\hat s$ is read at those fixed vertices; $L_{\mathrm{CV}}$ penalizes endpoint-value errors and $L_{\mathrm{pers}}$ penalizes the error in the predicted persistence $\hat p=|\hat d-\hat b|$. Gradients flow only through ordinary scalar-field values. Because no diagram of the prediction is computed or matched, this is a differentiable proxy for persistence agreement, not an optimization of a persistence-diagram distance.}
\label{fig:objective-schematic}
\end{figure*}

%% file: tables/tab_headline.tex
Bottleneck $d_B$ & 3.124 & 3.288 & \textbf{2.150} & $-$5.2\% & $+$31.2\% & $+$34.6\% & 151/168 & 156/168\\
$W_{2,\infty}$ & 19.15 & 21.06 & \textbf{14.33} & $-$10.0\% & $+$25.2\% & $+$32.0\% & 167/168 & 168/168\\
$W_{2,2}$ & 24.55 & 27.51 & \textbf{17.83} & $-$12.0\% & $+$27.4\% & $+$35.2\% & 166/168 & 168/168\\
Merge tree & 5.868 & 6.012 & \textbf{5.657} & $-$2.5\% & $+$3.6\% & $+$5.9\% & 98/168 & 124/168\\
\midrule
All three PD lower & & & & & & & 150/168 & 156/168\\
All three PD and MT lower & & & & & & & 91/168 & 119/168\\

%% file: tables/tab_all_candidates.tex
Pretrained CNN & 3.124 & 19.15 & 24.55 & 5.868 & \cellcolor{improve!11}$+$5.0 & \cellcolor{improve!15}$+$9.1 & \cellcolor{improve!17}$+$10.7 & \cellcolor{improve!8}$+$2.4 & 13/13/13/7\\
Pretrained GAN & 5.059 & 14.68 & \textbf{17.46} & 8.348 & \cellcolor{degrade!50}$-$53.9 & \cellcolor{improve!39}$+$30.3 & \cellcolor{improve!46}$+$36.5 & \cellcolor{degrade!49}$-$38.9 & 18/4/1/18\\
\midrule
Standard fine-tuning control & 3.288 & 21.06 & 27.51 & 6.012 & -- & -- & -- & -- & 17/18/18/12\\
\midrule
+ Speed & 3.262 & 21.01 & 27.43 & 6.000 & \cellcolor{improve!6}$+$0.8 & \cellcolor{improve!5}$+$0.2 & \cellcolor{improve!5}$+$0.3 & \cellcolor{improve!5}$+$0.2 & 15/16/16/10\\
+ Level-set & 3.270 & 21.03 & 27.46 & 6.008 & \cellcolor{improve!6}$+$0.5 & \cellcolor{improve!5}$+$0.2 & \cellcolor{improve!5}$+$0.2 & \cellcolor{improve!5}$+$0.1 & 16/17/17/11\\
+ Speed + Level-set & 3.188 & 20.82 & 27.14 & 5.944 & \cellcolor{improve!8}$+$3.0 & \cellcolor{improve!6}$+$1.2 & \cellcolor{improve!7}$+$1.3 & \cellcolor{improve!6}$+$1.1 & 14/15/15/8\\
+ Gradient & 2.544 & 15.14 & 18.69 & 6.056 & \cellcolor{improve!30}$+$22.6 & \cellcolor{improve!37}$+$28.1 & \cellcolor{improve!41}$+$32.0 & \cellcolor{degrade!6}$-$0.7 & 8/7/7/13\\
+ Speed + Gradient & 2.681 & 15.86 & 19.68 & 6.290 & \cellcolor{improve!26}$+$18.4 & \cellcolor{improve!33}$+$24.7 & \cellcolor{improve!37}$+$28.5 & \cellcolor{degrade!10}$-$4.6 & 10/10/10/17\\
+ Gradient + Level-set & 2.632 & 15.62 & 19.32 & 6.200 & \cellcolor{improve!27}$+$19.9 & \cellcolor{improve!34}$+$25.9 & \cellcolor{improve!38}$+$29.8 & \cellcolor{degrade!9}$-$3.1 & 9/9/9/16\\
+ Speed + Gradient + Level-set & 2.698 & 16.01 & 19.87 & 6.161 & \cellcolor{improve!25}$+$17.9 & \cellcolor{improve!32}$+$24.0 & \cellcolor{improve!36}$+$27.8 & \cellcolor{degrade!8}$-$2.5 & 11/12/12/15\\
\midrule
+ Speed + Gradient + Level-set + Local-max & 2.495 & 15.87 & 19.71 & 6.080 & \cellcolor{improve!32}$+$24.1 & \cellcolor{improve!33}$+$24.7 & \cellcolor{improve!37}$+$28.4 & \cellcolor{degrade!6}$-$1.1 & 7/11/11/14\\
+ Local-max & 2.817 & 20.17 & 26.25 & 5.690 & \cellcolor{improve!21}$+$14.3 & \cellcolor{improve!10}$+$4.2 & \cellcolor{improve!10}$+$4.6 & \cellcolor{improve!11}$+$5.4 & 12/14/14/6\\
+ Gradient + Local-max & 2.377 & 15.05 & 18.56 & 5.984 & \cellcolor{improve!36}$+$27.7 & \cellcolor{improve!37}$+$28.5 & \cellcolor{improve!42}$+$32.5 & \cellcolor{improve!6}$+$0.5 & 6/6/6/9\\
\midrule
+ Fixed-pair & 2.237 & 15.17 & 19.10 & \textbf{5.594} & \cellcolor{improve!41}$+$32.0 & \cellcolor{improve!36}$+$28.0 & \cellcolor{improve!39}$+$30.6 & \cellcolor{improve!13}$+$7.0 & 5/8/8/1\\
+ Speed + Gradient + Level-set + Fixed-pair & 2.169 & 14.48 & 18.04 & 5.677 & \cellcolor{improve!43}$+$34.0 & \cellcolor{improve!40}$+$31.3 & \cellcolor{improve!44}$+$34.4 & \cellcolor{improve!11}$+$5.6 & 3/3/4/5\\
+ Speed + Gradient + Level-set + Local-max + Fixed-pair & \underline{2.151} & 14.70 & 18.37 & 5.663 & \cellcolor{improve!44}$+$34.6 & \cellcolor{improve!39}$+$30.2 & \cellcolor{improve!42}$+$33.2 & \cellcolor{improve!12}$+$5.8 & 2/5/5/3\\
\textbf{+ Gradient + Fixed-pair (final)} & \textbf{2.150} & \underline{14.33} & 17.83 & \underline{5.657} & \cellcolor{improve!44}$+$34.6 & \cellcolor{improve!41}$+$32.0 & \cellcolor{improve!45}$+$35.2 & \cellcolor{improve!12}$+$5.9 & 1/2/3/2\\
+ Gradient + Level-set + Fixed-pair & 2.176 & \textbf{14.30} & \underline{17.76} & 5.674 & \cellcolor{improve!43}$+$33.8 & \cellcolor{improve!41}$+$32.1 & \cellcolor{improve!45}$+$35.4 & \cellcolor{improve!11}$+$5.6 & 4/1/2/4\\

%% file: tables/tab_factorial.tex
\multicolumn{5}{l}{\emph{(a) Speed $\times$ gradient $\times$ level-set: all eight combinations}}\\
\quad Speed & $-$0.024 & $-$0.212 & $-$0.283 & $-$0.030\\
\quad \textbf{Gradient} & $+$0.613 & $+$5.322 & $+$7.993 & $-$0.186\\
\quad Level-set & $-$0.004 & $-$0.098 & $-$0.119 & $+$0.011\\
\quad Speed$\times$gradient & $-$0.078 & $-$0.343 & $-$0.480 & $-$0.068\\
\quad Speed$\times$level-set & $+$0.032 & $+$0.121 & $+$0.170 & $+$0.081\\
\quad Gradient$\times$level-set & $-$0.049 & $-$0.211 & $-$0.290 & $-$0.019\\
\quad Three-way & $+$0.004 & $+$0.042 & $+$0.052 & $+$0.055\\
\midrule
\multicolumn{5}{l}{\emph{(b) Local-max $\times$ fixed-pair, added to speed + gradient + level-set}}\\
\quad Local-max & $+$0.111 & $-$0.042 & $-$0.084 & $+$0.048\\
\quad \textbf{Fixed-pair} & $+$0.436 & $+$1.348 & $+$1.580 & $+$0.451\\
\quad Local-max$\times$fixed-pair & $-$0.093 & $-$0.183 & $-$0.243 & $-$0.033\\
\midrule
\multicolumn{5}{l}{\emph{(c) Level-set $\times$ fixed-pair, added to gradient}}\\
\quad Level-set & $-$0.057 & $-$0.221 & $-$0.281 & $-$0.081\\
\quad \textbf{Fixed-pair} & $+$0.425 & $+$1.069 & $+$1.211 & $+$0.462\\
\quad Level-set$\times$fixed-pair & $+$0.031 & $+$0.251 & $+$0.349 & $+$0.063\\

%% file: tables/tab_medians.tex
Pretrained CNN & 2.913 & 18.816 & 24.057 & 5.712\\
Pretrained GAN & 4.196 & 13.518 & 15.835 & 7.587\\
Standard fine-tuning control & 3.188 & 21.290 & 27.841 & 5.972\\
+ Speed & 3.189 & 21.259 & 27.866 & 5.991\\
+ Level-set & 3.192 & 21.280 & 27.872 & 5.970\\
+ Speed + Level-set & 3.141 & 21.101 & 27.582 & 5.913\\
+ Gradient & 2.343 & 14.324 & 17.630 & 5.915\\
+ Speed + Gradient & 2.524 & 15.114 & 18.692 & 6.203\\
+ Gradient + Level-set & 2.473 & 14.878 & 18.331 & 6.058\\
+ Speed + Gradient + Level-set & 2.531 & 15.595 & 19.328 & 6.041\\
+ Speed + Gradient + Level-set + Local-max & 2.342 & 15.148 & 18.755 & 5.996\\
+ Local-max & 2.619 & 20.356 & 26.613 & 5.661\\
+ Fixed-pair & 2.050 & 15.259 & 18.728 & 5.502\\
+ Speed + Gradient + Level-set + Fixed-pair & 1.956 & 14.304 & 17.707 & 5.594\\
+ Speed + Gradient + Level-set + Local-max + Fixed-pair & 1.969 & 14.609 & 18.310 & 5.543\\
\textbf{+ Gradient + Fixed-pair (final)} & 1.934 & 14.122 & 17.531 & 5.519\\
+ Gradient + Level-set + Fixed-pair & 1.979 & 14.137 & 17.479 & 5.546\\
+ Gradient + Local-max & 2.231 & 14.424 & 17.855 & 5.926\\

%% file: tables/tab_wins.tex
Pretrained CNN & -- & -- & -- & -- & -- & -- & 102 & 146 & 151 & 104 & 93 & 67\\
Pretrained GAN & 39 & 155 & 158 & 20 & 39 & 16 & 43 & 167 & 168 & 21 & 43 & 16\\
Standard fine-tuning control & 62 & 22 & 17 & 64 & 11 & 7 & -- & -- & -- & -- & -- & --\\
+ Speed & 65 & 20 & 17 & 61 & 11 & 6 & 102 & 129 & 133 & 95 & 80 & 49\\
+ Level-set & 62 & 22 & 18 & 64 & 11 & 5 & 103 & 120 & 127 & 89 & 67 & 38\\
+ Speed + Level-set & 70 & 29 & 25 & 66 & 16 & 10 & 129 & 168 & 168 & 116 & 129 & 89\\
+ Gradient & 133 & 161 & 162 & 66 & 131 & 61 & 162 & 168 & 168 & 91 & 162 & 89\\
+ Speed + Gradient & 122 & 158 & 160 & 42 & 119 & 34 & 159 & 168 & 168 & 52 & 159 & 49\\
+ Gradient + Level-set & 126 & 158 & 162 & 41 & 123 & 36 & 161 & 168 & 168 & 63 & 161 & 60\\
+ Speed + Gradient + Level-set & 117 & 157 & 160 & 40 & 114 & 35 & 162 & 168 & 168 & 67 & 162 & 64\\
+ Speed + Gradient + Level-set + Local-max & 137 & 158 & 160 & 52 & 133 & 45 & 161 & 168 & 168 & 76 & 161 & 73\\
+ Local-max & 105 & 57 & 49 & 94 & 41 & 34 & 151 & 168 & 168 & 147 & 151 & 133\\
+ Fixed-pair & 148 & 165 & 166 & 104 & 148 & 100 & 156 & 168 & 168 & 136 & 156 & 132\\
+ Speed + Gradient + Level-set + Fixed-pair & 148 & 166 & 166 & 90 & 147 & 86 & 158 & 168 & 168 & 121 & 158 & 120\\
+ Speed + Gradient + Level-set + Local-max + Fixed-pair & 154 & 166 & 166 & 94 & 153 & 89 & 157 & 168 & 168 & 122 & 157 & 119\\
\textbf{+ Gradient + Fixed-pair (final)} & 151 & 167 & 166 & 98 & 150 & 91 & 156 & 168 & 168 & 124 & 156 & 119\\
+ Gradient + Level-set + Fixed-pair & 148 & 167 & 166 & 102 & 147 & 94 & 156 & 168 & 168 & 122 & 156 & 119\\
+ Gradient + Local-max & 139 & 161 & 161 & 61 & 138 & 55 & 163 & 168 & 168 & 90 & 163 & 88\\

%% file: tables/tab_contrasts.tex
Speed, added to control & $+$0.80\% & $+$0.25\% & $+$0.29\% & $+$0.20\%\\
Level-set, added to control & $+$0.54\% & $+$0.16\% & $+$0.19\% & $+$0.07\%\\
Gradient, added to control & $+$22.63\% & $+$28.10\% & $+$32.05\% & $-$0.73\%\\
Local-max, added to control & $+$14.31\% & $+$4.25\% & $+$4.59\% & $+$5.36\%\\
Fixed-pair, added to control & $+$31.97\% & $+$27.99\% & $+$30.57\% & $+$6.95\%\\
Local-max, added to S+G+L & $+$7.55\% & $+$0.88\% & $+$0.80\% & $+$1.31\%\\
Fixed-pair, added to S+G+L & $+$19.61\% & $+$9.56\% & $+$9.18\% & $+$7.85\%\\
Fixed-pair, added to S+G+L+M & $+$13.76\% & $+$7.34\% & $+$6.78\% & $+$6.87\%\\
Local-max, added to S+G+L+FP & $+$0.82\% & $-$1.55\% & $-$1.81\% & $+$0.26\%\\
Local-max, added to G & $+$6.56\% & $+$0.62\% & $+$0.72\% & $+$1.19\%\\
Fixed-pair, added to G (gives final) & $+$15.49\% & $+$5.40\% & $+$4.61\% & $+$6.60\%\\
Level-set, added to final G+FP & $-$1.24\% & $+$0.21\% & $+$0.38\% & $-$0.31\%\\
Gradient, added to FP (gives final) & $+$3.89\% & $+$5.55\% & $+$6.64\% & $-$1.12\%\\
Remove S and L from S+G+L+FP (gives final) & $+$0.89\% & $+$1.05\% & $+$1.17\% & $+$0.37\%\\